\documentclass[10pt,twocolumn,letterpaper]{article}
\usepackage[pagenumbers]{cvpr}

\usepackage{graphicx}
\usepackage{amsmath}
\usepackage{amssymb}
\usepackage{booktabs}
\usepackage[dvipsnames]{xcolor}
\usepackage{xr-hyper}
\usepackage{xr}
\usepackage{multicol}
\usepackage{multirow}
\usepackage{makecell}
\usepackage{pifont}
\usepackage{mathtools}
\usepackage{placeins}
\usepackage{etoc}

\usepackage{algorithm}
\usepackage{algpseudocodex}

\usepackage[accsupp]{axessibility} 

\usepackage[subtle,mathspacing=normal,wordspacing=normal,bibbreaks=normal,paragraphs=normal]{savetrees}

\definecolor{cvprblue}{rgb}{0.21,0.49,0.74}
\usepackage[pagebackref,breaklinks,colorlinks,citecolor=cvprblue]{hyperref}

\usepackage[capitalize]{cleveref}
\crefname{section}{Sec.}{Secs.}
\Crefname{section}{Section}{Sections}
\Crefname{table}{Table}{Tables}
\crefname{table}{Tab.}{Tabs.}

\newcommand{\tightcaption}[1]{\vspace{-0.072in}\caption{#1}\vspace{-0.2in}}

\newcommand{\cmark}{\ding{51}}
\newcommand{\xmark}{\ding{55}}

\makeatletter
\newcommand*{\addFileDependency}[1]{
\typeout{(#1)}
\@addtofilelist{#1}
\IfFileExists{#1}{}{\typeout{No file #1.}}
}\makeatother

\newcommand\rurl[1]{
  \href{https://#1}{\nolinkurl{#1}}
}

\newcommand{\vx}{\mathbf{x}}

\newcommand{\cx}{\mathcal{X}}

\newcommand{\vy}{\mathbf{y}}

\usepackage{bm}

\DeclareMathOperator*{\argmin}{argmin}
\DeclareMathOperator*{\argmax}{argmax}

\DeclarePairedDelimiter{\abs}{\lvert}{\rvert}
\DeclarePairedDelimiter{\norm}{\lVert}{\rVert}

\DeclarePairedDelimiter{\paren}{(}{)}
\DeclarePairedDelimiter{\curly}{\{}{\}}

\newcommand{\Prob}[1]{\text{Pr}\left\{#1 \right\}}

\newcommand{\tildeNice}{{\raise.17ex\hbox{$\scriptstyle\sim$}}}
\newcommand{\greaterNice}{{\raise.17ex\hbox{$\scriptstyle >$}}}
\newcommand{\approxNice}{{\raise.15ex\hbox{$\scriptstyle\approx$}}}
\newcommand{\ggNice}{{\raise.15ex\hbox{$\scriptstyle >> $}}}

\DeclareCaptionLabelFormat{andfigure}{#1~#2  \&  \figurename~\thefigure}
\DeclareCaptionLabelFormat{andtable}{#1~#2  \&  \tablename~\thetable}

\def\paperID{5218} 
\def\confName{CVPR}
\def\confYear{2024}

\author{
  Varun Sundar$^{\dagger} \footnote{Equal Contribution}$\\
  {\tt\small vsundar4@wisc.edu}\and
  Matthew Dutson $^{\dagger*}$\\
  {\tt\small dutson@wisc.edu}\and
  Andrei Ardelean$^{\ddagger}$\\
  {\tt\small a.ardelean@epfl.ch}\and
  Claudio Bruschini$^\S$ \quad Edoardo Charbon$^\S$\\
  {\tt\small \{claudio.bruschini,}
  {\tt\small edoardo.charbon\}@epfl.ch}\and
  Mohit Gupta$^{\dagger}$\\
  {\tt\small mohitg@cs.wisc.edu}\and
  {$^\dagger$University of Wisconsin-Madison}\quad
  {$^\ddagger$NovoViz}\quad
  {$^\S$École Polytechnique Fédérale de Lausanne}\\
  \framebox{\small{\rurl{wisionlab.com/project/generalized-event-cameras/}}}}

\title{Generalized Event Cameras}

\begin{document}

\setlength\abovedisplayskip{5.7pt}
\setlength\belowdisplayskip{7pt}
\setlength\abovedisplayshortskip{0pt}
\setlength\belowdisplayshortskip{4pt}

\maketitle

\etocdepthtag.toc{mtchapter}
\etocsettagdepth{mtchapter}{subsection}
\etocsettagdepth{mtappendix}{none}

\renewcommand*{\thefootnote}{$\ast$}
\setcounter{footnote}{1}
\footnotetext{denotes equal contribution.}

\renewcommand*{\thefootnote}{\arabic{footnote}}
\setcounter{footnote}{0}

\begin{abstract}
Event cameras capture the world at high time resolution and with minimal bandwidth requirements.
However, event streams, which only encode changes in brightness, do not contain sufficient scene information to support a wide variety of downstream tasks.
In this work, we design generalized event cameras that inherently preserve scene intensity in a bandwidth-efficient manner.
We generalize event cameras in terms of when an event is generated and what information is transmitted.
To implement our designs, we turn to single-photon sensors that provide digital access to individual photon detections; this modality gives us the flexibility to realize a rich space of generalized event cameras.
Our single-photon event cameras are capable of high-speed, high-fidelity imaging at low readout rates.
Consequently, these event cameras can support plug-and-play downstream inference, without capturing new event datasets or designing specialized event-vision models.
As a practical implication, our designs, which involve lightweight and near-sensor-compatible computations, provide a way to use single-photon sensors without exorbitant bandwidth costs.
\end{abstract}
\vspace{-0.2in}

\section{Introduction}
\label{sec:intro}

\begin{figure*}[t]
    \centering
    \includegraphics[width=\textwidth]{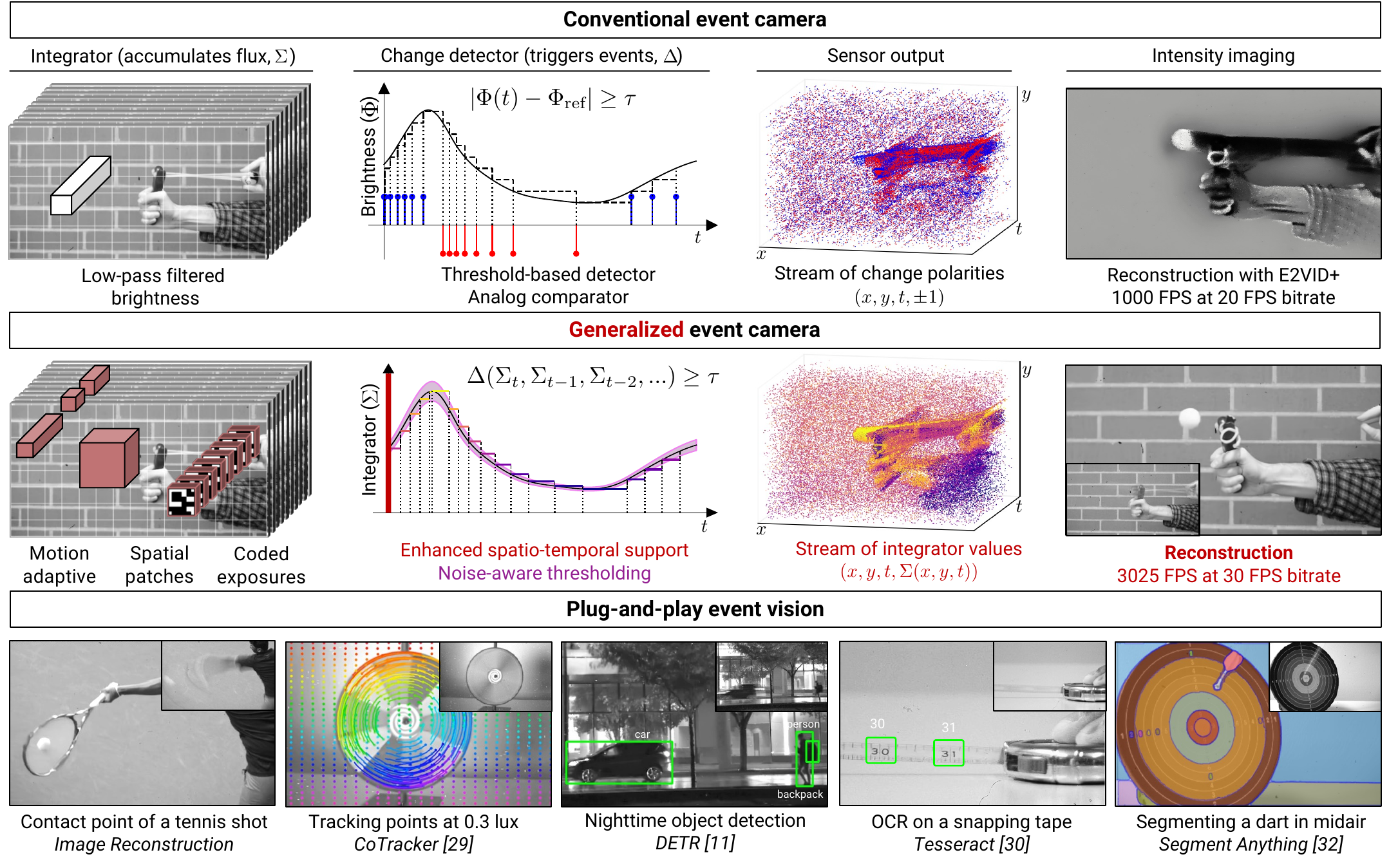}
    \vspace{-0.2in}
    \tightcaption{\textbf{Generalized event cameras.} 
    \textit{(top)} Event cameras generate outputs in response to abrupt changes in scene intensity.
    We describe this as a combination of a low-pass integrator and a threshold-based change detector.
    \textit{(middle)} We generalize the space of event cameras by designing \textit{integrators} that capture rich intensity information, and more reliable \textit{change detectors} that utilize larger spatiotemporal contexts and noise-aware thresholding (\cref{sec:bitplane_events,sec:chunk_events,sec:coded_events}). 
    Unlike existing events, our generalized event streams inherently preserve scene intensity, \eg, this ping-pong ball slingshotted against a brick wall backdrop.
    \textit{(bottom)} Generalized event cameras enable high-fidelity bandwidth-efficient imaging: providing $3025$~FPS reconstructions with a readout equivalent to a $30$ FPS camera.
    Consequently, generalized events facilitate \emph{plug-and-play inference} on a multitude of tasks in challenging scenarios (insets depict the extent of motion over $30$ ms).}
    \label{fig:first-figure}
    \vspace{0.1in}
\end{figure*}
\nocite{karaev2023cotracker}
\nocite{kay2007tesseract}

Event cameras~\cite{lichtsteiner200364x64,atis_2011,berner2013240x180} sense the world at high speeds, providing visual information with \emph{minimal bandwidth and power}.
They achieve this by transmitting only changes in scene brightness, when significant ``events'' occur. However, there is a cost. Raw event data, a sparse stream of binary values, does not hold sufficient information to be used directly with mainstream vision algorithms.
Therefore, while event cameras have been successful at certain tasks (\eg, object tracking~\cite{gehrig2020eklt,li2021tracking}, obstacle avoidance~\cite{he2021fast,bisulco2021fast,sanket2020evdodgenet}, and high-speed odometry~\cite{censi2014low,Zhu_2017_CVPR,hidalgo-carrio_2022_cvpr}), they are not widely deployed as general-purpose vision sensors, and often need to be supplemented with conventional cameras~\cite{gehrig2020eklt,hidalgo-carrio_2022_cvpr,GaoPAMI}. These limitations are holding back this otherwise powerful technology.

Is it possible to realize the promise of event cameras, \ie, high temporal resolution at low bandwidth, \textit{while preserving} rich scene intensity information? To realize these seemingly conflicting goals, we propose a novel family of \textit{generalized event cameras}.
We conceptualize a space of event cameras along two key axes (\cref{fig:first-figure}~\textit{(top)}): (a) ``when to transmit information,'' formalized as a change detection procedure $\Delta$; and (b) ``what information to transmit,'' characterized by an integrator $\Sigma$ that encodes incident flux. Existing event cameras represent one operating point in this $(\Sigma, \Delta)$ space. Our key observation is that by exploring this space and considering new $(\Sigma,\Delta)$ combinations, we can design event cameras that preserve scene intensity.
We propose more general integrators, \eg, that represent flux according to motion levels, that span spatial patches, or that employ temporal coding (\cref{fig:first-figure} \textit{(middle)}). We also introduce robust change detectors that better distinguish motion from noise, by considering increased spatiotemporal contexts and modeling noise in the sensor measurements.

Despite their conceptual appeal, physically implementing generalized event cameras is a challenge. This is because the requisite computations must be performed \emph{at the sensor} to achieve the desired bandwidth reductions. For example, existing event cameras perform simple integration and thresholding operations via analog in-pixel circuitry. However, more general $(\Sigma, \Delta)$ combinations are not always amenable to analog implementations; even feasible designs might require years of hardware iteration and production scaling. To build physical realizations of generalized event cameras, we leverage an emerging sensor technology: single-photon avalanche diodes (SPADs) that provide \textit{digital access} to photon detections at extremely high frame rates ($\tildeNice 100$~kHz). This allows us to compose arbitrary \textit{software-level} signal transformations, such as those required by generalized event cameras. Further, we are not locked to a particular event camera design and can realize multiple configurations with the same sensor.\smallskip

\noindent \textbf{Implications: extreme, bandwidth-efficient vision.} Generalized event cameras support high-speed, high-quality image reconstruction, but at low bandwidths quintessential of current event cameras. For example, \cref{fig:first-figure}~\textit{(middle, bottom)} shows reconstructions at $3025$~FPS that have an effective readout of a $30$~FPS frame-based camera. Further, our methods have strong low-light performance due to the SPAD's single-photon sensitivity. As we show in \cref{fig:first-figure}~\textit{(bottom)}, preserving scene intensity facilitates plug-and-play inference in challenging scenarios, with state-of-the-art vision algorithms. Critically, this does not require retraining vision models or curating dedicated datasets, which is a significant challenge for unconventional imagers.
This plug-and-play capability is vital to realizing universal event vision that retains the benefits of current event cameras.\smallskip

\noindent \textbf{Scope.} We consider \emph{full-stack} event perception: we conceptualize a novel space of event cameras, provide relevant single-photon algorithms, analyze their imaging capabilities and rate-distortion trade-offs, and show on-chip feasibility. We demonstrate imaging capabilities in \cref{sec:experiments-videography,sec:experiments-applications} using the SwissSPAD2 array~\cite{ulku512512spad2019}, and show viable implementations of our algorithms for UltraPhase~\cite{ardelean2023computational}, a recent single-photon compute platform. All of these are critical to unlocking the promise of event cameras. However, our objective is not to develop an integrated system that incorporates all these components; this paper merely takes the first steps toward that goal. 

\section{Related Work}
\label{sec:related_work}

\noindent \textbf{Event camera designs.} Perhaps most widespread is the DVS event camera \cite{lichtsteiner200364x64}, 
where each pixel generates an event in response to measured changes in (log) intensity.
The DAVIS event camera \cite{davis240,berner2013240x180} couples DVS pixels with conventional CMOS pixels, providing access to image frames. However, the frames lack the dynamic range of DVS events.
A recent design, Celex-V~\cite{HuangCelexV}, provides log-intensity frames using an external trigger. ATIS~\cite{atis_2011}, a less prevalent design, features asynchronous intensity events, but its sophisticated circuitry reduces pixel fill factor.
The above designs are based on analog processing; we instead design event cameras on \textit{digital} photon detections.\smallskip

\noindent \textbf{Intensity imaging with event cameras.} Several approaches have been explored to obtain images from events, including Poisson solvers~\cite{barua2016direct}, manifold regularization~\cite{munda2018real}, assuming knowledge of camera motion \cite{kim2014simultaneous,cook2011} or optical flow~\cite{zhang2023formulating}, and learning-based methods~\cite{rebecq19pami,scheerlinck_2020_wacv,Zou_2021_CVPR,suUnsupICIP,Paredes-Valles_2021_CVPR}. However, because events often lack sufficient scene information,
they are often supplemented by conventional frames~\cite{scheerlinck2018continuous,brandli2014real,shedligeri2019photorealistic,Pan_2019_CVPR}, either from sensors such as DAVIS or using a multi-camera setup. Fusing events and frames presents challenges due to potential spatiotemporal misalignment and discrepancies in imaging modalities.
Even when these challenges are overcome, we show that fusion methods produce lower fidelity than our proposed generalized event cameras.\smallskip

\noindent\textbf{Passive single-photon imaging.} In the past few years, SPADs have found compelling passive imaging applications; this includes high-dynamic range imaging~\cite{ingleHighFluxPassive2019,inglePassiveInterPhotonImaging2021,Liu_2022_WACV,namiki2022imaging}, motion deblurring~\cite{Iwabuchi2021,ma_quanta_2020,Ma_2023_WACV,laurenzis2023single,laurenzis2022comparison}, high-speed tracking~\cite{gyongy2018single}, and ultra wide-band videography~\cite{Wei_2023_ICCV}. A particularly relevant method is proposed by \citet{seets_2021_wacv}, which uses flux changepoint estimation to perform burst photography on single-photon sequences. This approach uses flux changepoints to estimate motion, then integrates along spatiotemporal motion trajectories to circumvent the noise-blur tradeoff. This spatiotemporal integration allows for high-quality reconstructions under challenging lighting and motion conditions. In contrast, our paper emphasizes changepoint estimation as a means to compress single-photon data. Further, since we aim to run our proposed techniques near sensor, where there are limited memory and compute capabilities, we focus on online changepoint estimation that processes photon detections in a single pass.

The fine granularity of passive single-photon acquisition makes it possible to emulate a diverse set of imaging modalities~\cite{sundar_sodacam_2023}, including event cameras, via post-capture processing. In this work, we go beyond emulating existing event cameras and design alternate event cameras that preserve high-fidelity intensity information.

\section{What is an Event Camera?}
\label{sec:technical_commentary}

The defining characteristic of event cameras is that they transmit information \textit{selectively}, in response to changes in scene content. This selectivity allows event cameras to encode scene information at high time resolutions required to capture scene dynamics, \textit{without} proportionately high bandwidth. This is in contrast to frame-based cameras, where readout occurs at fixed intervals.

We characterize event cameras in terms of two axes: \textit{what} the camera transmits and \textit{when} it transmits. 
As a concrete example, consider existing event cameras. They trigger events (``when to transmit'') based on a fixed threshold:
\begin{equation}
\label{eq:event-generation}
    |\Phi(\vx, t) - \Phi_\text{ref}(\vx)| \geq \tau,
\end{equation}
where $\Phi(\vx, t)$ is a flux estimate\footnote{Event cameras such as the DVS~\cite{davis240} measure a temporally low-pass filtered estimate of log-flux; we absorb this into $\Phi(\vx, t)$ for brevity.} at pixel $\vx$ and time $t$, and $\tau$ is the threshold. $\Phi_\text{ref}(\vx)$ is a previously-recorded reference, set to $\Phi(\vx, t)$ whenever an event is triggered. Each event consists of a packet
\begin{equation}
\label{eq:binary-event}
    \paren*{\vx, t, \text{sign}\paren*{\Phi(\vx, t) - \Phi_\text{ref}(\vx)}}
\end{equation}
that encodes the polarity of the change (``what to transmit'').

Event polarities, although adequate for some applications, do not retain sufficient information to support a general set of computer vision tasks. A stream of event polarities is an extremely lossy representation. Notably, it only defines scene intensity up to an initial unknown reference value, and it does not encode any information in regions not producing events, \ie, regions with little or no motion.

\begin{figure}[t]
    \centering
    \includegraphics{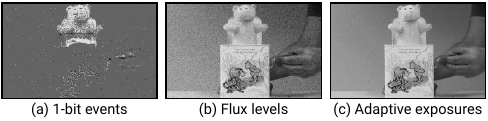}
    \tightcaption{\textbf{Altering ``what to transmit.''} 
    \textit{(a)} We sum the events generated by a jack-in-the-box toy as it springs up. This sum 
    gives a lossy encoding of brightness changes in dynamic regions.
    \textit{(b)} Transmitting levels instead of changes helps recover details in static regions. 
    \textit{(c)} Adaptive exposures, which accumulate flux between consecutive events, provide substantial noise reduction.}
    \label{fig:telescoping}
\end{figure}

Our key observation is that existing event cameras represent just one operating point in a broader space of \textit{generalized event cameras}, which is defined by two axes: ``what to transmit'' and ``when to transmit.'' By considering alternate points in this space, we can design event cameras that preserve high-fidelity scene intensity. This enables plug-and-play inference with a host of algorithms developed by the mainstream vision community.

We begin with a conceptual exploration of this generalized space, before describing its physical implementation.

\begin{table*}[t]
    \centering
    \resizebox{\textwidth}{!}{\begin{tabular}{@{}l | c c | c c c c @{}}
        \toprule
        Event camera &
        \makecell[c]{Integrator ($\Sigma$)}    &
        \makecell[c]{Change detector ($\Delta$)}  &
        \makecell[c]{Event packets}  & 
        \makecell[c]{Min. latency}  & 
        \makecell[c]{Intensity info.?} &
        \makecell[c]{Low-light perf.} \\ 
        \midrule
        Existing (DVS~\cite{lichtsteiner200364x64}) & 
        logarithmic & comparator & binary & $10^{-6}$ to $10^{-5}$s 
        & \xmark & poor\\
        \cref{sec:bitplane_events} & adaptive exposure & Bayesian change detector \cite{alami2020restarted} & scalar & $10^{-5}$s 
        & \cmark & good\\
        \cref{sec:chunk_events} & adaptive exposure & variance-aware differences & patches & $10^{-4}$s 
        & \cmark & good\\
        \cref{sec:coded_events} & coded exposure & Binomial confidence interval & vector & $10^{-3}$s 
        & \cmark & good\\
     \bottomrule
    \end{tabular}}
    \vspace{-0.1in}
    \caption{\textbf{Summary of generalized event cameras.} 
    Our designs integrate photon detections ($\Sigma$) and detect scene-content changes ($\Delta$) in distinct ways.
    We compare our designs to existing DVS event cameras based on their event streams, latencies, and intensity-preserving nature. 
    While providing a direct power comparison to DVS is difficult, we compare the power characteristics among our designs in \cref{sec:experiments-ultraphase}.
    }
    \vspace{-0.15in}
    \label{tab:method_overview}
\end{table*}

\smallskip
\noindent \textbf{Generalizing ``what to transmit.''} As a first step, we can modify the event camera such that it transmits $n$-bit values instead of one-bit change polarities. When an event is triggered, we send the current value of $\Phi(\vx, t)$; if a pixel triggers no events, we transmit $\Phi$ during the final readout. As we show in \cref{fig:telescoping}~(b), this simple change allows us to recover scene intensity, even in static regions.
It is important to note that, while the transmitted quantity differs from conventional events, we retain the defining feature of an event camera: selective transmission based on scene dynamics (where we transmit according to \cref{eq:event-generation}). Thus, the readout remains decoupled from the time resolution.

If $\Phi(\vx, t)$ were a perfect estimate of scene intensity, then the changes thus far would suffice. However, $\Phi$ is fundamentally noisy: to capture high-speed changes, $\Phi$ must encompass a shorter duration, which leads to higher noise. This is a manifestation of the classical noise-blur tradeoff.

To address this problem, we introduce the abstraction of an \textit{integrator} (or $\Sigma$), that defines how we accumulate incident flux and, in turn, what we transmit.
Ideally, we want the integrator to \textit{adapt} to scene dynamics, \ie, accumulate over longer durations when there is less motion, and vice versa. 
We observe that event generation, which is based on scene dynamics, can be used to formulate an adaptive integrator. Specifically, we propose an integrator $\Sigma_\text{cuml}$ that computes the cumulative flux since the last event:
\begin{equation}
\label{eq:cuml-integrator}
    \Sigma_\text{cuml}(\vx, t) = \int_{T_0}^{t} \Phi(\vx, s) ds,
\end{equation}
where $T_0$ is the time of the last event. When an event is triggered at time $T_1$, we communicate the value\footnote{We can either transmit values of $\Sigma_\text{cuml}$ or differences (changes) to $\Sigma_\text{cuml}$; we treat this as an implementation detail here.} of $\Sigma_\text{cuml}(\vx, T_1)$, which we interpret as the intensity throughout $[T_0, T_1]$. This approach yields a piece-wise constant time series, with segments delimited by events. We note that a similar idea, of virtual exposures beginning and ending with change points, was also explored in \citet{seets_2021_wacv} as part of a motion-adaptive deblurring pipeline. Adaptive exposures significantly reduce noise while preserving dynamic scene content, as we show in \cref{fig:telescoping}~\textit{(c)}.

\smallskip
\noindent \textbf{Generalizing ``when to transmit.''} The success of the adaptive integrator crucially depends on the reliability of events; for example, triggering false events in static regions causes unnecessary noise. We refer to the event-generation procedure as the \textit{change detector}, denoted by $\Delta$. Current event cameras detect changes by applying a fixed threshold to measured intensity differences (\cref{eq:event-generation}).
This method has two key limitations: it only considers the value of $\Phi$ at pixel location $\vx$ and time $t$ and is not attuned to the stochasticity in $\Phi$. 

We design more robust change detectors that (1) leverage enhanced spatiotemporal contexts, and (2) incorporate noise awareness, either explicitly by tuning thresholds, or implicitly by modulating the detector's behavior. Specifically, we improve reliability by using temporal forecasters (\cref{sec:bitplane_events}), by leveraging correlated changes in patches (\cref{sec:chunk_events}), or by exploiting integrator statistics  (\cref{sec:coded_events}). 

\smallskip
\noindent \textbf{Realizing generalized event cameras.} The critical detail remaining is how we implement our proposed designs in practice. We need direct access to flux estimates at a high time resolution.
Conventional high-speed cameras can provide such access,
however, they incur substantial per-frame read noise ($\tildeNice 20$--$40$$e^-$~\cite{igual2019photographic}) that grows with frame rate~\cite{readout_noise}.

We turn to an emerging class of single-photon sensors, single-photon avalanche diodes (SPADs~\cite{rochas2003single}), that has witnessed dramatic improvements in device practicality and key sensor characteristics (\eg, array sizes and fill factors) in recent years~\cite{ulku512512spad2019,morimoto_megapixel_2020}. 
SPADs can operate at extremely high speeds ($\tildeNice 100$~kHz) without incurring per-frame read noise. 
Each $\Phi(\vx, t)$ measured by a SPAD is limited only by the fundamental stochasticity of photon arrivals (shot noise). This allows SPADs to provide high timing resolution without a substantial noise penalty. In the next section, we describe the image formation model of SPADs and provide single-photon implementations of our designs.
\section{Single-Photon Generalized Event Cameras}
\label{sec:sigma-delta-instances}

A SPAD array can operate as a high-speed photon detector, producing binary frames as output. Each binary value indicates whether at least one photon was detected during an exposure. The SPAD output response, $\Phi(\vx, t)$, can be modeled as a Bernoulli random variable, with
\begin{equation}
\label{eq:bernoulli}
    P \left( \Phi(\vx, t) = 1 \right) = 1 - e^{-N(\vx, t)},
\end{equation}
where $N(\vx, t)$ is the average number of photo-electrons during an exposure, including any spurious detections. The \textit{inherently digital} SPAD response allows us to compute software-level transformations on the signal $\Phi(\vx, t)$, including operations that may be challenging to realize via analog processing. These transformations can be readily reconfigured, which permits a spectrum of event camera designs, not just one particular choice. However, there is one consideration: our designs should be lightweight and computable on chip. As we show in \cref{sec:experiments-ultraphase}, this is vital to implementing generalized event cameras without the practical costs associated with reading off raw SPAD outputs.

We now describe a set of SPAD-based event cameras (summarized in \cref{tab:method_overview}), beginning with the adaptive exposure method from the previous section.

\smallskip
\noindent\textbf{Adaptive-exposure event camera.} 
We obtain a SPAD implementation of the adaptive exposure described in \cref{eq:cuml-integrator} by replacing the integral with a sum over photons:
\begin{equation}
\label{eq:cuml-sum}
    \Sigma_\text{cuml}(\vx, t) = \sum_{s = T_0}^{t} \Phi(\vx, s).
\end{equation}
To generate events, we can use a threshold-based change detector (\cref{eq:event-generation}). Differences between individual binary values are not sufficiently informative; therefore, we apply \cref{eq:event-generation} to an exponential moving average (EMA) computed on $\Phi$. We call this an ``adaptive-EMA'' event camera.

\subsection{Bayesian Change Detector}
\label{sec:bitplane_events}

\begin{figure}[t]
    \centering
    \includegraphics{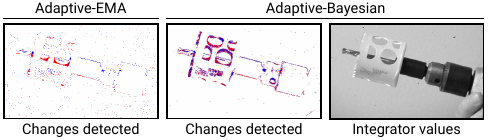}
    \tightcaption{\textbf{Bayesian- vs. EMA-based change detection}. 
    \textit{(left)} A fixed-threshold change detector (used in adaptive-EMA) makes it difficult to segment low-contrast changes.
    \textit{(center)} The Bayesian formulation attunes to the stochasticity in incident flux and can detect fine-grained changes such as the corners of the hole saw bit; \textit{(right)} as a result, the integrator captures the rotational dynamics.} 
    \label{fig:ema-bocpd}
\end{figure}

A fixed-threshold change detector such as \cref{eq:event-generation} does not account for the SPAD's image formation model; it uses the same threshold irrespective of the underlying variance in photon detections. As a result, such a detector may fail to detect changes in low-contrast regions without producing a large number of false-positive detections (see \cref{fig:ema-bocpd} \textit{(left)}).

In this section, we consider a Bayesian change detector, BOCPD~\cite{adams2007bayesian}, that is tailored to the Bernoulli statistics of photon detections.
BOCPD uses a series of \textit{forecasters} to estimate the likelihood of an abrupt change.
At each time step, a new forecaster $\nu_t$ is initialized as a recurrence of previous forecasters, and existing forecasters are updated:
\begin{equation}
\label{eq:forecaster-update}
    \nu_t = (1 - \gamma) \sum_{s = 1}^{t - 1} l_s \nu_s, \qquad \nu_s \gets \gamma l_s \nu_s \enspace \forall \; s < t,
\end{equation}
where $\gamma \in [0, 1]$  is the sensitivity of the change detector, with larger $\gamma$ resulting in more frequent detections. $l_s$ is the predictive likelihood of each forecaster, which we compute by tracking two values per forecaster, $\alpha_s$ and $\beta_s$, that correspond to the parameters of a Beta prior. For a new forecaster, these values are initialized to $1$ each, reflecting a uniform prior. Existing $(\alpha_s, \beta_s)$, $\forall\, s < t$, are updated as
\begin{equation}
\label{eq:posterior-update}
    \alpha_s \gets \alpha_s + \Phi(\vx, t), \quad \beta_s  \gets \beta_s + 1 - \Phi(\vx, t).
\end{equation}
$l_s$ is given by $\alpha_s / (\alpha_s + \beta_s)$ if $\Phi(\vx, t) = 1$, and $\beta_s / (\alpha_s + \beta_s)$ otherwise. 
An event is triggered if the highest-value forecaster does not correspond to $T_0$, the timestamp of the last event; mathematically, if ${\text{argmax}}_t\, \nu_t \neq T_0$.

To make BOCPD viable in memory-constrained scenarios, we apply extreme pruning by retaining only the three highest-value forecasters~\cite{wang2021online}. We also incorporate restarts, deleting previous forecasters when a change is detected~\cite{alami2020restarted}.

Compared to an EMA-based change detector, the Bayesian approach more reliably triggers events in response to scene changes while better filtering out stochastic variations caused by photon noise---which we show in \cref{fig:ema-bocpd}.

\subsection{Spatiotemporal Chunk Events}
\label{sec:chunk_events}

\cref{sec:bitplane_events} leverages an expanded \textit{temporal} context for change detection; however, it treats each pixel independently and does not exploit \textit{spatial} information.
In this section, we devise an event camera with enhanced spatial context that operates on small patches, \eg, of $4 \times 4$ pixels.
It is difficult to derive efficient Bayesian change detectors for multivariate time series; thus, we adopt a \textit{model-free} approach that does not explicitly parameterize the patch distribution. To afford computational breathing room for more expensive patch-wise operations, we employ temporal chunking. That is, we average $\Phi(\vx, t)$ over a small number of binary frames (\eg, 32 binary frames) instead of operating on individual binary frames; generally, this averaging does not induce perceptible blur.

\begin{figure}
    \centering
    \includegraphics{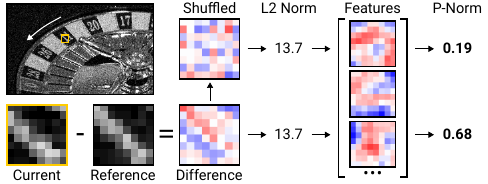}
    \tightcaption{\textbf{Spatiotemporal chunk events.} We evaluate the difference between the current chunk and a stored reference in a learned linear-feature space. Unlike  the L2 norm, which is permutation-invariant, the feature-space norm is sensitive to spatial structure. Randomly shuffling the pixel values reduces the transform-domain norm (the shuffled patch has a more ``noise-like'' structure).}
    \label{fig:patch_method}
\end{figure}

Let vector $\bm{\Phi}_\text{chunk}(\vy, t)$ represent the chunk-wise average of photon detections at patch location $\vy$. Let vector $\bm{\Sigma}_\text{patch}(\vy, t)$ be an integrator representing the cumulative mean since the last event, 
but excluding $\bm{\Phi}_\text{chunk}$.
We want to estimate whether $\bm{\Phi}_\text{chunk}$ belongs to the same distribution as $\bm{\Sigma}_\text{patch}$. We do so with a lightweight approach, that computes the distance between $\bm{\Phi}_\text{chunk}$ and $\bm{\Sigma}_\text{patch}$ in the linear feature space of matrix $\mathbf{P}$.
As we show in \cref{fig:patch_method}, linear features allow us to capture spatial structure within a patch. Geometrically, $\mathbf{P}$ induces a hyperellipsoidal decision boundary, in contrast to the spherical boundary of the L2 norm.

This method generates an event whenever
\begin{equation}
\label{eq:patch-events}
    \norm{\mathbf{P}\paren{\tilde{\bm{\Phi}}_\text{chunk}(\vy, t) - \tilde{\bm{\Sigma}}_\text{patch}(\vy, t)}}_2 \geq \tau,
\end{equation}
where $\tau$ is the threshold. When there is no event, we extend the cumulative mean to include the current chunk. Before computing linear features, we normalize $\bm{\Phi}_\text{chunk}$ and $\bm{\Sigma}_\text{patch}$ element-wise according to the estimated variance in $\bm{\Phi}_\text{chunk} - \bm{\Sigma}_\text{patch}$; we annotate the normalized versions with a tilde. We estimate the variance based on the fact that, in a static patch, the elements of $\bm{\Phi}_\text{chunk}$ and $\bm{\Sigma}_\text{patch}$ are independent binomial random variables.

We train the matrix $\mathbf{P}$ on simulated SPAD data, generated from interpolated high-speed video. We apply backpropagation through time to minimize the MSE error of the transmitted patch values. To address the non-differentiability arising from the threshold, we employ surrogate gradients. Please see the supplementary material for complete details of this method.

\subsection{Coded-Exposure Events}
\label{sec:coded_events}

In this section, we design a generalized event camera by applying change detection to coded exposures~\cite{raskar2006coded,P2C2,hitomi2011VCS}, which capture temporal variations by multiplexing photon detections over an integration window.
This is interesting in two aspects. First, we are designing event streams based on an modality not typically associated with event cameras. Second, we show that high-speed information can be obtained even when the change detector operates at a coarser time granularity.
Coded-exposure events provide somewhat lower fidelity than our designs in \cref{sec:bitplane_events,sec:chunk_events}, but are more compute- and power-efficient, owing to less frequent execution of the change detector.

At each pixel, we multiplex a temporal chunk of $T_\text{code}$ ($\tildeNice 256$--$512$) binary values with a set of $J$ ($\tildeNice 2$--$6$) codes $C^j(\vx, t) \,\forall\, 1\leq j \leq J$,
producing $J$ coded exposures
\begin{equation}
\label{eq:multi-bucket}
    \Sigma_{\text{coded}}^j(\vx, t) = \sum_{s=t - T_\text{code}}^{t} \Phi(\vx, s) C^j(\vx, s).
\end{equation}
The codes $C^j$ are chosen to be random, mutually orthogonal binary masks, each containing $T_\text{code} / \text{max}(2, J)$ ones~\cite{sundar_sodacam_2023}. 

We exploit the statistics of coded exposures to derive a change detector. Observe that in static regions, $\Sigma_\text{coded}^j(\vx, t)$ are independent and identically distributed (iid) binomial random variables.
Thus, we can expect them to lie within a binomial confidence interval of one another. If not, we assume the pixel is dynamic and generate an event. 
We trigger an event if $\Sigma_\text{coded}^j \notin \text{conf} (n, \hat{p})$ for any $j$.
Here, ``$\text{conf}$'' refers to a binomial confidence interval (\eg, Wilson's score), $n = T_\text{code} / J$ draws, and $\hat{p} = \sum_s \Phi(\vx, s) / T_\text{code}$ is the empirical success probability.

If a pixel is static, we store the sum of the $J$ coded exposures, which is a long exposure, denoted by $\Sigma_\text{long}$. If the pixel remains static across more than one temporal chunk, we extend $\Sigma_\text{long}$ to include the entire duration. Whereas, if the pixel is dynamic, we transmit $\curly{\Sigma^j_\text{coded}}$, as well as any previous static intensity encoded in $\Sigma_\text{long}$. 
Downstream, we can apply coded-exposure restoration techniques~\cite{wang_2023_cvpr,Yuan2022PnPADMM,Shedligeri_2021_WACV} to recover intensity frames from the coded measurements.

\section{Experimental Results}
\label{sec:experiments}

We demonstrate the capabilities of generalized event cameras using a SwissSPAD2 array~\cite{ulku512512spad2019} with resolution $512 \times 256$, which we use to capture one-bit frames at $96.8$~kHz.
We show the feasibility of our designs on {UltraPhase}~\cite{ardelean2023computational}, a recent single-photon computational platform (\cref{sec:experiments-ultraphase}).\smallskip

\noindent \textbf{Refinement model.} For each of our event cameras, we train a refinement model that mitigates artifacts arising from the asynchronous nature of events. This model takes a periodic frame-based sampling of integrator values and outputs a video reconstruction. The sampling rate is configurable; in practice, we set it $\tildeNice 16$--$64\times$ lower than the SPAD rate. We use a densely-connected residual architecture~\cite{wang_2023_cvpr}, trained on data generated by simulating photon detections on temporally interpolated~\cite{huang2022rife} high-speed videos from the XVFI dataset~\cite{Sim_2021_ICCV}. See the supplement for training details.

\begin{figure}[t]
    \centering
    \includegraphics{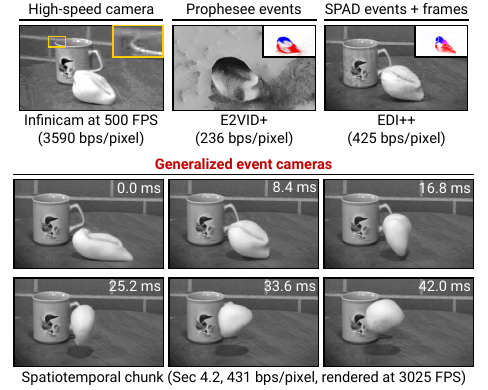}
    \tightcaption{\textbf{High-speed videography of a stress ball hurled at a coffee mug.} \textit{(top row)} This indoor scene is challenging for existing imaging systems, including: high-speed cameras (SNR-related artifacts), event cameras (poor restoration quality), and even hybrid event + frame techniques (reconstruction artifacts).
    \textit{(bottom rows)} In contrast, our generalized event cameras capture the stress ball's extensive deformations with high fidelity and an efficient readout.
    }
    \label{fig:high_speed_recons}
\end{figure}

\begin{figure*}[t]
    \centering
    \includegraphics{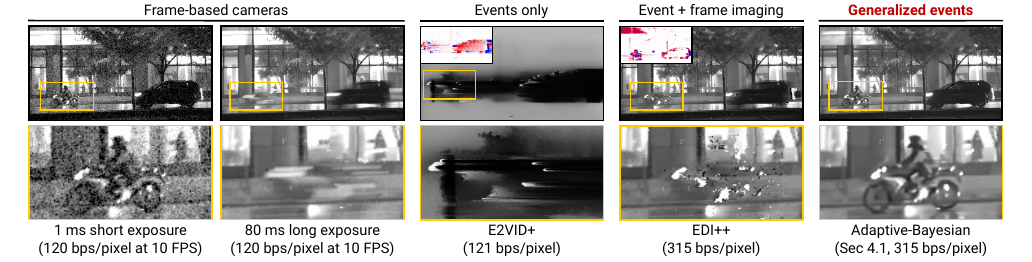}
    \vspace{-0.2in}
    \tightcaption{\textbf{Event imaging in urban nighttime ($7$ lux, sensor side).}
    \textit{(left to right)} Low-light conditions necessitate long exposures in frame-based cameras, resulting in unwanted motion blur. 
    The Prophesee EVK4 suffers from severe degradation in low light, causing E2VID+ to fail.
    Running EDI++ on perfectly aligned SPAD-frames and -events improves overall restoration quality but still gives failures on fast-moving objects. Our generalized events recover significantly more detail in low light, as seen in the inset of the motorcyclist.
    }
    \label{fig:low_light_recons}
\end{figure*}

\begin{figure*}
    \centering
    \vspace{0.1in}
    \includegraphics{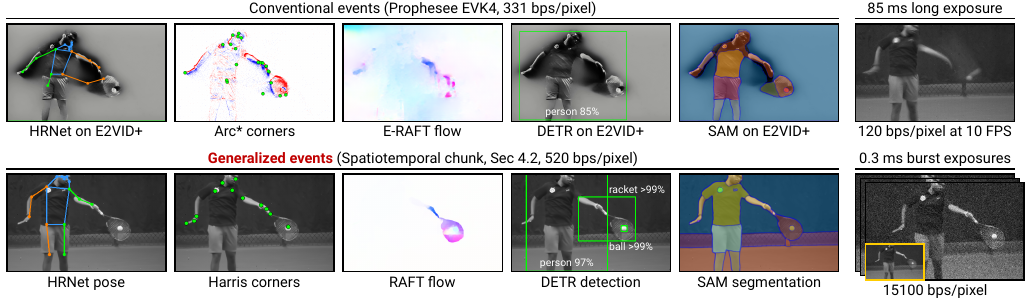}
    \vspace{-0.2in}
    \tightcaption{\textbf{Plug-and-play inference on a tennis scene.} 
    \textit{(top left)} Conventional events encode temporal-gradient polarities; this lossy representation limits performance on downstream tasks.
    \textit{(bottom left)} Generalized events encode rich scene-intensity information, with a readout comparable existing event cameras. They facilitate high-quality plug-and-play inference, without requiring dedicated algorithms. 
    \textit{(right)} Generalized event cameras give image quality comparable to burst photography techniques that have a much higher readout rate.
    }
    \vspace{0.05in}
    \label{fig:plug_and_play}
\end{figure*}

\subsection{Extreme Bandwidth-Efficient Videography}
\label{sec:experiments-videography}

\noindent \textbf{High-speed videography.} In \cref{fig:high_speed_recons}, we capture the dynamics of a deformable ball (a ``stress ball'') using a SPAD, a high-speed camera (Photron Infinicam) operated at $500$~FPS, and a commercial event camera (Prophesee EVK4). The high-speed camera suffers from low SNR due to read noise, which manifests as prominent artifacts after on-camera compression. Meanwhile, conventional events captured by Prophesee, when processed by ``intensity-from-events'' methods such as E2VID+~\cite{stoffregen2020reducing} fail to recover intensities reliably, especially in static regions. We also evaluate EDI~\cite{pantpami2022}, a hybrid event-frame method. We consider an idealized variant that operates on SPAD events (obtained via EMA thresholding), which gives perfect event-frame alignment and a precisely known event-generation model.
We refine the outputs of EDI using the same model as for our methods. We refer to this idealized, refined version of EDI as ``EDI++.'' While EDI++ recovers more detail than other baselines, there are considerable artifacts in its outputs.

Our method achieves high-quality reconstructions at $3025$~FPS ($96800 / 32$) that faithfully capture non-rigid deformations, with only $431$~bits per second per pixel (bps/pixel) readout, which is a $227\times$ compression ($96800 / 431$) of the raw SPAD capture. Viewed differently, for a $1$~MPixel array, we would obtain a bitrate of $431$~Mbps, implying that we can read off these $3025$ FPS reconstructions over USB 2.0 (which supports transfer speeds of up to $480$~Mbps).\smallskip

\noindent \textbf{Event imaging in low light.} \cref{fig:low_light_recons} compares the low-light performance of frame-based, event-based, and a generalized event camera on an urban night-time scene at $7$~lux (lux measured at the sensor). For frame-based cameras, a short exposure that preserves motion may be too noisy, while a long exposure can be severely blurred. The Prophesee's performance deteriorates in low light, resulting in blurred temporal gradients. EDI++, benefiting from the idealized SPAD-based implementation, can image this scene, but finer details like the motorcyclist are lost. Our generalized event cameras, on the other hand, provide reconstructions with minimal noise, blur, or artifacts---while retaining the bandwidth efficiency of event-based systems. The compression here is $307\times$ with respect to raw SPAD outputs.

\subsection{Plug-and-Play Inference}
\label{sec:experiments-applications}

Generalized event cameras preserve scene intensity, which enables plug-and-play event-based vision. We consider a tennis sequence (of $8196$~binary frames) containing a range of object speeds.
We evaluate a range of tasks: pose estimation (HRNet~\cite{sun2019hrnet}), corner detection~\cite{harris1988corner}, optical flow (RAFT~\cite{teed2020RAFT}), object detection (DETR~\cite{nicolas2020detr}), and segmentation (SAM~\cite{kirillov2023segment}). We compare against event-based methods applied to Prophesee events; we use Arc*~\cite{alzugaray2018arcstar} for corner detection and E-RAFT~\cite{gehrig2021raft} for optical flow.
For the remaining tasks, which do not have equivalent event methods, we run HRNet, DETR, and SAM on E2VID+ reconstructions.

As \cref{fig:plug_and_play}~\textit{(top left)} shows, traditional events are bandwidth efficient ($331$~bps/pixel), 
but do not provide sufficient information for successful inference.
Generalized events \textit{(bottom left)} have a modestly higher readout ($520$~bps/pixel), but support accurate inference without requiring dedicated algorithms. 
To provide context for these rates, we compare them against frame-based methods \textit{(right)}. A long exposure ($120$~bps/pixel) blurs out the racket. Burst methods~\cite{hasinoff_burst_2016} recover a sharp image from a stack of short exposures, but with a large readout of $15100$~bps/pixel.

\subsection{Rate-Distortion Analysis}
\label{sec:rate-distortion-analysis}

Each method in \cref{sec:sigma-delta-instances} features a sensitivity parameter that controls the sensor readout rate (event rate), which in turn influences image quality.
In this subsection, we evaluate the impact of readout on image quality (PSNR) by performing a rate-distortion analysis.
For ground truth, we use a set of YouTube-sourced high-speed videos captured by a Phantom Flex4k at $1000$~FPS; see the supplement for thumbnails and links. 
We upsample these videos to the SPAD's frame rate and then simulate $4096$ binary frames using the image formation model described in \cref{eq:bernoulli}.
When computing readout for our methods, we assume that events encode $10$-bit values and account for the header bits of each event packet.

\begin{figure}[t]
    \centering
    \includegraphics[width=\columnwidth]{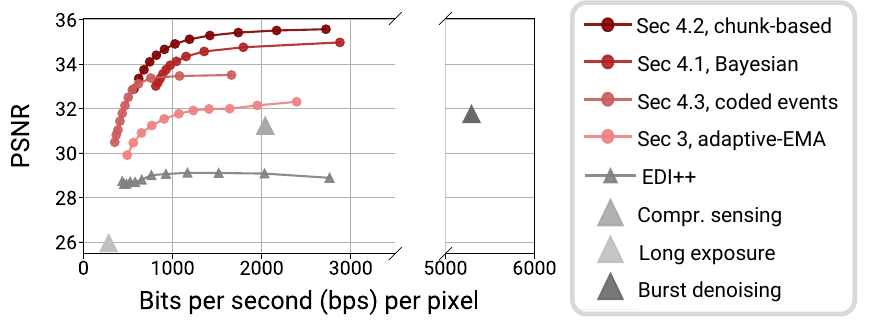}
    \vspace{-0.15in}
    \tightcaption{\textbf{Rate-distortion evaluation.} Our techniques feature a tunable parameter that controls the output event rate. Generalized events offer a $4$--$8$ dB improvement in PSNR over EDI++ (at the same readout), and can compress raw photon data by $80\times$.}
    \label{fig:rate_distortion}
\end{figure}

As baselines, we consider EDI++, a long exposure, compressive sensing with 8-bucket masks, and burst denoising~\cite{hasinoff_burst_2016} using $32$ short exposures. 
As \cref{fig:rate_distortion} shows, generalized event cameras provide a pronounced $4$--$8$~dB PSNR improvement over baseline methods.
Further, our methods can compress the raw SPAD response by around $80\times$ before a noticeable drop-off in PSNR is observed.

Among our methods, the spatiotemporal chunk approach of \cref{sec:chunk_events} gives the best PSNR, followed by the Bayesian method (\cref{sec:bitplane_events}) and coded-exposure events (\cref{sec:coded_events}). That said, all methods are fairly similar in terms of rate-distortion (\eg, all three give comparable results for the scenes in \cref{sec:experiments-videography,sec:experiments-applications}).
The methods are better distinguished by their practical characteristics.
The Bayesian method gives single-photon temporal resolution; however, as we show in \cref{sec:experiments-ultraphase}, it is the most expensive to compute on-chip. The chunk-based method occupies a middle ground in terms of latency and cost.
Coded-exposure events have the highest latency---events are generated only every $\tildeNice256$--$512$ binary frames---but the lowest on-chip cost. This provides an end user the flexibility to choose from the space of generalized event cameras based on the latency requirements and the compute constraints of the target application.

\subsection{On-Chip Feasibility and Validation}
\label{sec:experiments-ultraphase}

A critical limitation of single-photon sensors is the exorbitant bandwidth and power costs involved in reading off raw photon detections. However, the lightweight nature of our event camera designs allows us to sidestep this limitation by performing computations on-chip.
We demonstrate that our methods are feasible on UltraPhase~\cite{ardelean2023computational}, a SPAD compute platform. UltraPhase consists of 3$\times$6 compute cores, each of which is associated with 4$\times$4 pixels.

We implement our methods for UltraPhase using custom assembly code. Some methods require minor modifications due to instruction-set limitations; see the supplement for details. We process $2500$ SPAD frames from the tennis sequence used in \cref{sec:experiments-applications}, cropped to the UltraPhase array size of $12 \times 24$ pixels. We determine the number of cycles required to execute the assembly code and estimate the chip's power consumption and readout bandwidth.

\begin{figure}
    \centering
    \includegraphics{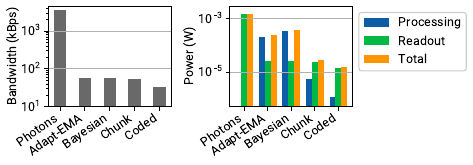}
    \vspace{-0.05in}
    \tightcaption{\textbf{On-chip compatibility.} We validate the feasibility of our approach on UltraPhase~\cite{ardelean2023computational}, a computational SPAD imager. Compared to reading out raw photon data, all of our approaches give marked reductions in both bandwidth \textit{(left)} and power \textit{(right)}.} 
    \label{fig:ultraphase}
\end{figure}

All our proposed methods run comfortably within the chip's compute budget of $4202$ cycles per binary frame and its memory limit of $4$~Kibit per core. As seen in \cref{fig:ultraphase}, compared to raw photon-detection readout, our techniques reduce both bandwidth and power costs by over two orders of magnitude. 
The coded-exposure method is particularly efficient; on most binary frames, it only requires multiplying a binary code with incident photon detections. 
Our proof-of-concept evaluation may pave the way for future near-sensor implementations of generalized event cameras, which with advances in chip-to-chip communication, could involve a dedicated ``photon processing unit'', similar to a camera image signal processor (ISP).

\section{Limitations and Discussion}
\label{sec:discussion}

\begin{figure}[t]
    \centering
    \vspace{0.15in}
    \includegraphics{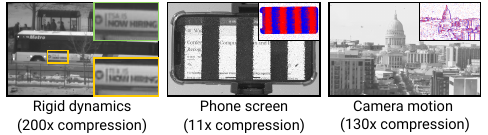}
    \tightcaption{\textbf{Limitations and failure modes.} \textit{(left)} Our reconstructions (yellow inset) on dynamic scenes with rigid objects can be inferior to burst photography (green inset). \textit{(center)} Modulated light sources, such as this phone screen, can trigger a deluge of events (change points shown in the inset). \textit{(right)} Rapid camera motion can result in an event rate divergent from scene dynamics.}
    \label{fig:limitations}
\end{figure}

Generalized events push the frontiers of event-based imaging; however, some scenarios lead to sub-optimal performance. As seen in \cref{fig:limitations}~\textit{(left)}, if the scene dynamics is entirely comprised of rigid motion, burst photography~\cite{ma_quanta_2020} gives better image quality, albeit with much higher readout. (\textit{middle}) Similar to current event cameras, modulated light sources trigger unwanted events that reduce bandwidth savings. However, it may be possible to ignore some of these events, perhaps by modeling the lighting variations~\cite{Sheinin_2017_CVPR}.\smallskip

\noindent\textbf{Ego-motion events.} Camera motion can trigger events in static regions, although our methods still yield substantial compression ($130\times$ over SPAD outputs, \cref{fig:limitations}~\textit{right}). We analyze the impact of ego-motion on bandwidth savings further in the supplement. However, single-photon cameras can emulate sensor motion by integrating flux along alternate spatiotemporal trajectories~\cite{sundar_sodacam_2023}. We can imagine a generalized event camera that is ``ego-motion compensated,'' by computing events along a suitable trajectory.\smallskip

\noindent \textbf{Photon-stream compression.} SPADs generate a torrent of data---\eg, $12.5$~GBps for a MPixel array at $100$~kHz---that can easily overwhelm data interfaces. Generalized event cameras reduce readout by around two orders of magnitude, by decoupling readout from the SPAD's frame rate and instead basing it on scene dynamics. This could pave the way for practical, high-resolution single-photon sensors.\smallskip

\section{Acknowledgments}

This research was supported in parts by NSF CAREER award 1943149, NSF award CNS-2107060, and the Swiss National Science Foundation grant 200021\_166289. 
We thank Paul Mos for providing us access to SwissSPAD2 acquisition software.
We also thank Trevor Seets for valuable discussions during the early stages of our research.

{
    \small 
    \setlength{\bibsep}{0pt}
    \bibliographystyle{ieeenat_fullname}
    \bibliography{references/changepoints,references/event_cameras,references/experiments,references/spad,references/main}
}
\clearpage

\onecolumn
\appendix
\begin{center}
{
    \Large
    \textbf{Supplementary Material}
}
\end{center}

\etocdepthtag.toc{mtappendix}
\etocsettagdepth{mtchapter}{none}
\etocsettagdepth{mtappendix}{subsection}
\tableofcontents
\clearpage

\section{Generalized Event Algorithms}
\label{supp_sec:algos}

\subsection{Overview}
\label{supp_sec:algos-overview}

\begin{figure}[htp]
    \centering
    \includegraphics[width=\textwidth]{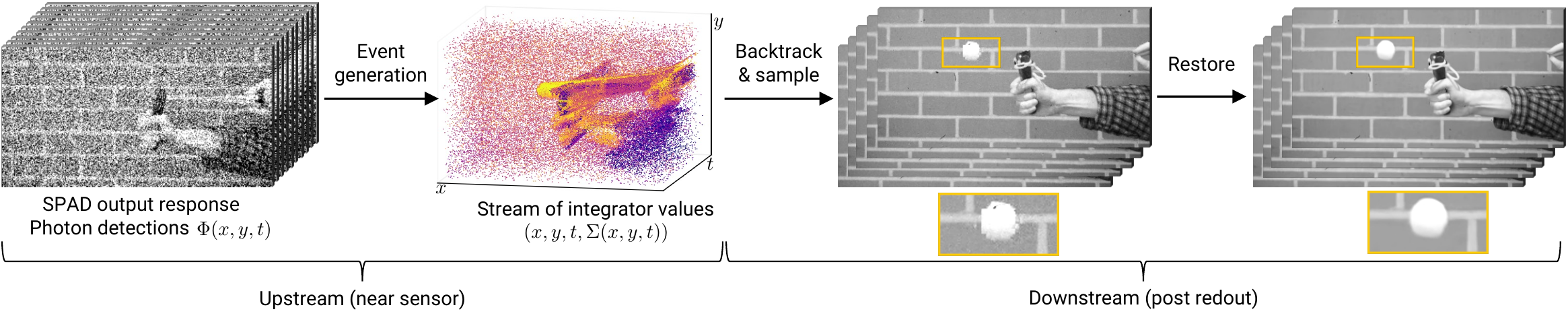}
    \vspace{-0.3in}
    \caption{\textbf{Algorithmic overview of generalized event cameras.} Our algorithms take as input the SPAD's response, $\Phi(\vx, t)$, and output a stream of integrator values, \ie, a stream of tuples $(\vx, t, \Sigma(\vx, t))$, where $\Sigma$ represents the integrator's value. After sensor readout, we perform \textit{backtracking}, which takes the value of the integrator as the flux estimate between the current and previous event timestamps at location $\vx$. We then sample these flux estimates at any time $t$, providing a stack of backtracked frames. Notice the rich scene information present in these backtracked frames. To alleviate artifacts (shown in the insets here) arising from the pixel-wise independent emission of events, we perform video restoration to recover high-quality outputs.}
    \label{fig:supp_overview}
\end{figure}

\cref{fig:supp_overview} shows an overview of recovering scene intensity in a bandwidth-efficient manner using generalized event cameras. This process begins from the SPAD's output response, $\Phi(\vx, t)$ at pixel location $\vx$ and (discrete) time $t$, and culminates in a high-fidelity, high-temporal resolution video reconstruction. In what follows, we go over the salient steps before laying down specific algorithms (in \cref{sec:supp_adapt_ema,sec:supp_adapt_bayesian,sec:supp_chunk,sec:supp_coded}) and detailing any algorithm-specific modifications to these steps.

\paragraph{Event generation.} Each of our algorithms (\cref{sec:sigma-delta-instances,sec:bitplane_events,sec:chunk_events,sec:coded_events}) takes as input the high-speed binary frames produced by the SPAD, $\Phi(\vx, t)$, and processes it in an online manner---without any buffering of photon detections. The output of our algorithms is an asynchronous (each pixel or patch can emit events independently) spatiotemporal stream of event packets.

\paragraph{Event packets.} For our methods, we assume events are sparsely encoded using a coordinate list (COO) format. In other words, we represent each event as a tuple $(\vx, t, \Lambda)$, where $\vx$ is the spatial location, $t$ is the time, and $\Lambda$ is the event payload---which is typically the integrator's value ($\Sigma$), except in the case of coded-exposure events where we transmit a set of integrator values. For the adaptive-EMA and Bayesian methods, $\Lambda = \Sigma_\text{cuml}$, the adaptive integrator. For the spatiotemporal chunk method, $\Lambda = \bm{\Sigma}_\text{patch}$, the patch-wise cumulative mean (a vector). For the coded method, $\Lambda = (\Sigma_\text{long}, \{\Sigma_\text{coded}^j\})$, the long exposure and most recent coded exposures (note that we can exclude $\Sigma_\text{long}$ if this is the change detector's first timestep or if we produced an event on the timestep immediately preceding this one).

\paragraph{Backtracking.} When an event is emitted at time $t=T_1$, the integrator's value, $\Sigma$, is taken to be the flux estimate (or representation) between $T_1$ and the previous event emission time ($T_0$)---if this is the first event emission, we assume $T_0=0$. Thus, from the spatiotemporal event stream, we can construct a piece-wise constant representation of the incident flux, by traversing the event stream backward in time; we term this process as \textit{backtracking}. 

\paragraph{Sampling.} After backtracking, we obtain a spatiotemporal cube of intensity estimates. We can now sample this cube discretely to obtain one or more frame-based samples. The main motivation for sampling the intensity cube, rather than processing it in its entirety, is that existing video restoration models are not capable of inference on very long video sequences (most models infer on $32$--$64$ video frames at a time). In this work, we consider a very simple sampling strategy: temporally uniform sampling. There can be, however, more sophisticated ways to sample, potentially based on the rate of events across time~\cite{seets_2021_wacv}---we do not explore these more sophisticated variants in this paper.

\paragraph{Restoration.} Having obtained a frame-based sampling (video representation) of our backtracked cube, we can now process it using video restoration techniques. The purpose of video restoration here is to remove artifacts arising from the independence of events between pixels (or spatial patches). For instance, neighboring pixels may fire events at different times, which tends to produce jagged artifacts around motion boundaries. We show these asynchronous artifacts (and their removal) in the insets of \cref{fig:supp_overview}.

\clearpage

\subsection{Adaptive-EMA Event Camera}
\label{sec:supp_adapt_ema}

\begin{figure}[htp]
    \centering
    \includegraphics[width=\textwidth]{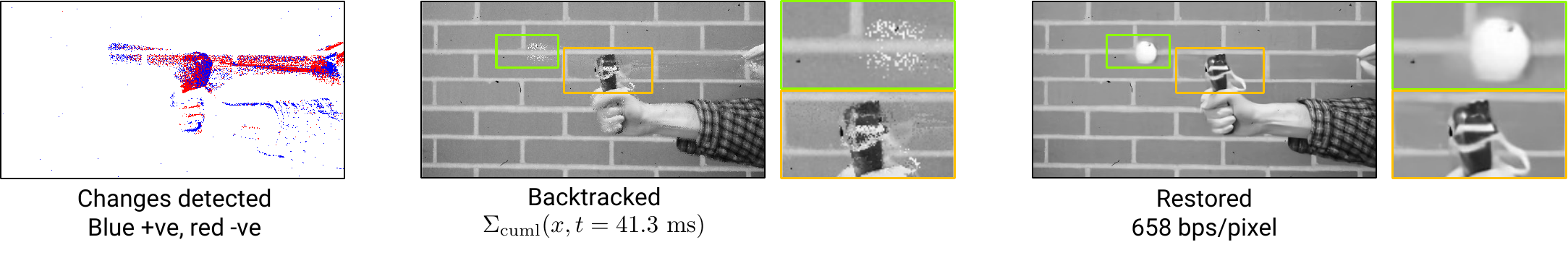}
    \vspace{-0.3in}
    \caption{\textbf{Intermediate outputs and recovered intensity from adaptive-EMA}. \textit{(left)} We show the changes detected on the slingshot sequence, across $4000$ binary frames. \textit{(middle)} The backtracked frame bears noticeable artifacts, \textit{(right)} however this is substantially reduced by merging information from multiple backtracked frames using a video restoration model. Improved reconstructions can be obtained using our more sophisticated methods (\cref{sec:bitplane_events,sec:chunk_events,sec:coded_events}) involving more robust change detection procedures.}
    \label{fig:adapt-ema-overview}
\end{figure}

In \cref{alg:adapt-ema}, we describe our simplest variant of the adaptive exposure technique, which was introduced in \cref{sec:technical_commentary} and further specified in \cref{sec:sigma-delta-instances}. This technique uses a fixed threshold applied to exponential moving averages (EMAs) to determine changes in scene intensity. Such a choice of change detector is motivated by early works in change-point detection that have utilized exponential moving averages \cite{roberts2000control}. We show intermediate and final outputs of adaptive-EMA on the slingshot sequence in \cref{fig:adapt-ema-overview}.

\begin{algorithm}
\caption{Adaptive-EMA Event Camera. Typical values of $\tau$ and $\gamma$ are in the range $0.4$--$0.5$, and $0.95$--$0.98$ respectively.}\label{alg:adapt-ema}
\begin{algorithmic}[1]
\Require {SPAD response, $\Phi(\vx, t)$ \\
Contrast threshold, $\tau$\\
Exponential smoothing factor, $\gamma$\\
Pixel locations, $\mathcal{X}$\\
Initial time-interval, $T_0$, for computing reference moving average\\
Total bit-planes, $T$}
\Ensure {Event-stream $E$ that contains packets of $(\vx, t, \text{cumulative mean})$}
\Function{GenerateEvents}{$\Phi(\vx, t)$, $\tau$, $\gamma$, $T_0$}
\For{$\vx \in \cx$}
\State Reference moving average, $\Sigma_\text{ref} \gets 0$
\State Current moving average, $\Sigma_\text{EMA} \gets 0$
\State Cumulative sum, $\Sigma_\text{cuml} \gets 0$
\State Cumulative counter, $n_\text{cuml} \gets 0$
\For{$t \in \{1, \ldots, T\}$}
\State $\Sigma_\text{EMA} \gets \gamma \Sigma_\text{EMA} + (1-\gamma) \Phi(\vx, t)$
\State $\Sigma_\text{cuml} \gets \Sigma_\text{cuml} + \Phi(\vx, t)$
\State $n_\text{CMA} \gets n_\text{CMA}+ 1$
\If{$t = T_0$}
    \State $\Sigma_\text{ref}(\vx) \gets \Sigma_\text{EMA}(\vx, t)$
\ElsIf{$T_0 < t < T$}
    \If{$\abs{\Sigma_\text{EMA} - \Sigma_\text{ref}} > \tau$}
        \State $E \gets (\vx, t, {\Sigma_\text{cuml}} / {n_\text{cuml}})$ \Comment{change detected}
        \State $\Sigma_\text{ref} \gets \Sigma_\text{cuml}$
        \State $\Sigma_\text{cuml} \gets 0$ \Comment{reset cumulative count}
        \State $n_\text{cuml} \gets 0$
    \EndIf
\ElsIf{$t= T$}
    \State $E \gets (\vx, t, {\Sigma_\text{cuml}(\vx, t)} / {n_\text{cuml}(\vx, t)})$ \Comment{flush out any pending updates}
\EndIf
\EndFor
\EndFor\\
\Return $E$
\EndFunction
\end{algorithmic}
\end{algorithm}

\clearpage

\subsection{Adaptive-Bayesian Event Camera}
\label{sec:supp_adapt_bayesian}

\begin{figure}[htp]
    \centering
    \includegraphics[width=\textwidth]{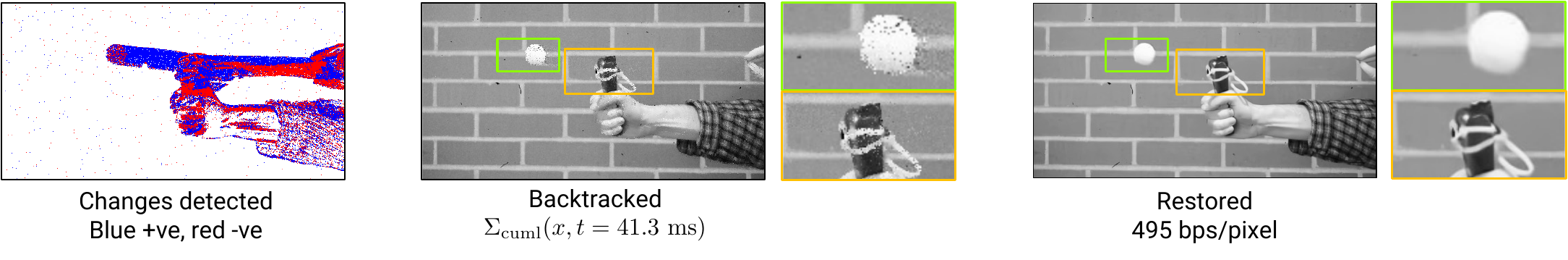}
    \vspace{-0.3in}
    \caption{\textbf{Intermediate outputs and recovered intensity from adaptive-Bayesian}. \textit{(left to right)} The change points detected by restarted-BOCPD are more informative that those detected by an EMA-based change detector. While there appears to be more change points here than in \cref{fig:adapt-ema-overview}, the changes are transmitted more parsimoniously over time---as a result, the overall readout is lower ($495$ bps/pixel for adaptive-Bayesian versus $658$ bps/pixel for adaptive-EMA). Using an improved change detector (BOCPD) also results in backtracked images that preserve more detail, and consequently a final reconstructed image that has significantly less blur (the ping-pong ball is better recovered here).}
    \label{fig:adapt-bocpd-overview}
\end{figure}

\begin{figure}[htp]
    \centering
    \includegraphics[width=\textwidth]{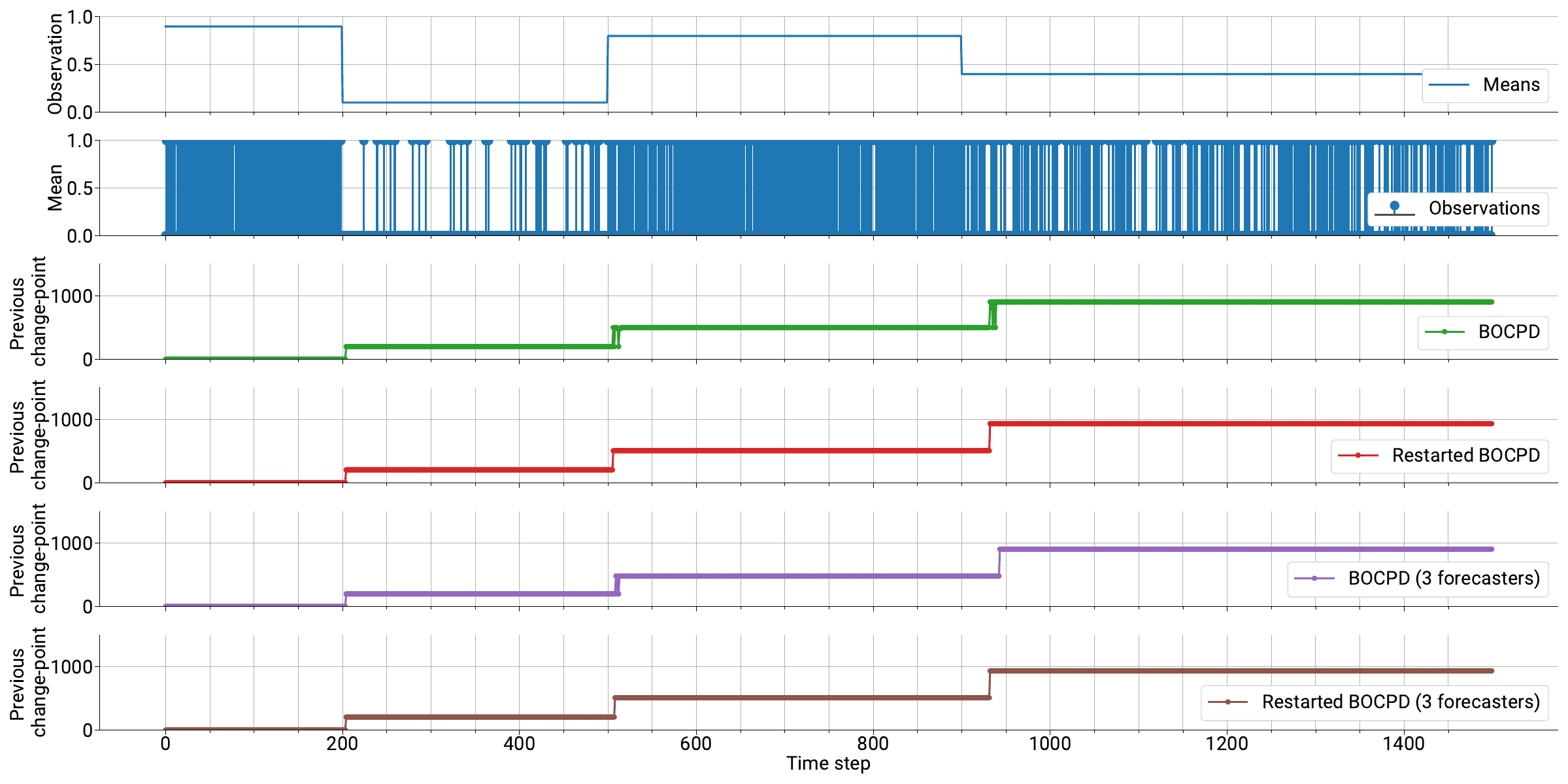}
    \caption{\textbf{Impact of restarts and pruning on change detection}. Using a (simple 1D) synthetic example, we show the impact of the restart strategy described by \citet{alami2020restarted} and extreme pruning (retaining just top-$3$ forecasters by value). \textit{(first two rows)} We consider a piece-wise stationary time series from which we draw Bernoulli samples. \textit{(third and fourth rows)} Incorporating restarts (and the pseudo distribution) helps reduce the detection delay. \textit{(last two rows)} Meanwhile, we do not see a substantial impact on the detector's performance when pruning the number of forecasters (for either BOCPD or restarted BOCPD). Plotted here are the values of the previous change-point timestep, as Bernoulli observations come in.}
    \label{fig:changepoint-example}
\end{figure}

In \cref{alg:adapt-bayesian}, we describe our generalized event camera (from \cref{sec:bitplane_events}), which uses (a variant of) restarted Bayesian online change detection (R-BOCPD~\cite{alami2020restarted}) as the per-pixel change detector. As mentioned in \cref{sec:bitplane_events}, our main modification is the use of \textit{extreme pruning}~\cite{wang2021online}: instead of storing one forecaster for each timestep (or per binary frame), we only retain the top-$K$ forecasters. We found that the performance of the change detector is not significantly impacted even with substantial pruning where we retain just the top-$3$ forecasters; see \cref{fig:changepoint-example}. We now describe a few variants of \cref{alg:adapt-bayesian}.

\paragraph{Restarts and pseudo-distribution array.} We can also consider the base variant of BOCPD~\cite{adams2007bayesian}, which does not involve restarts or an array of pseudo-distribution values. In this case, we can remove the arrays in line $7$ of \cref{alg:adapt-bayesian}, and initialize a new forecaster as $\nu^\prime = \gamma \sum_k \nu_k l_k$. Removing restarts allows us to skip lines $35$--$38$ in \cref{alg:adapt-bayesian}, and replace line $31$ by $\argmax_k \nu_k \neq T_0$, where $T_0$ is the timestamp of the previous event at the same pixel location $\vx$. We use this simplification for our on-chip implementation for UltraPhase~\cite{ardelean2023computational}.

\paragraph{Direct extensions to spatial patches.} While BOCPD is a per-pixel temporal change detector, there are simple ways to exploit the correlated changes observed in a patch---although these modifications should be seen as simple extensions and not a more principled approach, such as what we describe in \cref{sec:chunk_events}. To reduce detection delay, we can fire events whenever a changepoint is detected in \textit{any} pixel in a patch. While this may be preemptive for pixels where change has not yet been detected, being preemptive may not be detrimental, as a change in one pixel indicates that changes may soon be observed at other pixels in a patch.\\

We show intermediate and final outputs of adaptive-Bayesian on the slingshot sequence in \cref{fig:adapt-bocpd-overview}. Notice that the dynamics of the slingshot's elastic band and the propelled ping-pong ball are much better preserved in \cref{fig:adapt-bocpd-overview} (as compared to \cref{fig:adapt-ema-overview}), while entailing a reduced sensor readout as well.

\begin{algorithm}[htp]
\caption{Adaptive-Bayesian Event Camera. Typical ranges for the decay factor $\gamma$ are $[10^{-6}, 10^{-3}]$, with larger values resulting a more frequent change detection.}\label{alg:adapt-bayesian}
\begin{algorithmic}[1]
\Require {SPAD response, $\Phi(\vx, t)$ \\
Decay factor, $\gamma$\\
Number of forecasters, $K$\\
Pixel locations, $\mathcal{X}$\\
Total bit-planes, $T$}
\Ensure {Event-stream $E$ that contains packets of $(\vx, t, \text{cumulative mean})$}
\Function{GenerateEvents}{$\Phi(\vx, t)$, $\gamma$}
\For{$\vx \in \cx$}
\State Forecasters, $\nu_k$ with $1 \leq k \leq K$. 
\State \qquad $\nu_1 \gets 1$, $\nu_k \gets 0$ for all $k >1$.
\State Forecaster indices, $\mathcal{I}_k$ with $1 \leq k \leq K$. 
\State \qquad $\mathcal{I}_1 \gets 1$, $\mathcal{I}_k \gets 0$ for all $k >1$.
\State Pseudo-distribution, $\tilde{l}_k$ with $1 \leq k \leq K$. 
\State \qquad $\tilde{l}_1 \gets 1$, $\tilde{l}_k \gets 0$ for all $k >1$.
\State Alphas (of a Beta prior), $\alpha_k$ with $1 \leq k \leq K$. 
\State \qquad $\alpha_1 \gets 1$, $\alpha_k \gets 0$ for all $k >1$.
\State Betas (of a Beta prior), $\beta_k$ with $1 \leq k \leq K$. 
\State \qquad $\beta_1 \gets 1$, $\beta_k \gets 0$ for all $k >1$.
\State Cumulative sum, $\Sigma_\text{cuml} \gets 0$
\State Cumulative counter, $n_\text{cuml} \gets 0$
\For{$t \in \{1, \ldots, T\}$}
\State $\Sigma_\text{cuml} \gets \Sigma_\text{cuml}+ \Phi(\vx, t)$
\State $n_\text{CMA} \gets n_\text{CMA}+ 1$
\For{$k$ such that $\alpha_k, \beta_k >0$}
\Comment{compute predictive likelihoods}
\State Define $l_k = \alpha_k / (\alpha_k + \beta_k)$ if $\Phi(\vx, t) =1$
\State \quad $l_k = \beta_k / (\alpha_k + \beta_k)$ otherwise.
\Comment{update older forecasters}
\State $\nu_k \gets (1 - \gamma) \nu_k l_k $ 
\Comment{update priors}
\State $\alpha_k \gets \alpha_k + \Phi(\vx, t)$
\State $\beta_k \gets \beta_k + \Phi(\vx, t)$
\EndFor
\State $\nu^\prime = \gamma \sum_{k=1}^K \tilde{l}_k$ \Comment{new forecaster}
\If{$\nu^\prime > \min_k \nu_k$} \Comment{check if the new forecaster can replace an older one}
\State Denote $k_\text{min}= \argmin_k \nu_k$
\State $\tilde{l}_{k_\text{min}} \gets \nu^\prime$
\State $\nu_{k_\text{min}} \gets \nu^\prime$
\State $\mathcal{I}_{k_\text{min}} \gets t$
\State $\alpha_{k_\text{min}} \gets 1,\; \beta_{k_\text{min}} \gets 1$ \Comment{initialize uniform prior}
\EndIf
\If{$\argmax_k \nu_k \neq 1$} \Comment{change detected}
    \State $E \gets (\vx, t, {\Sigma_\text{cuml}} / {n_\text{cuml}})$ 
    \State $\Sigma_\text{cuml} \gets 0$     \Comment{reset cumulative count}
    \State $n_\text{cuml} \gets 0$
    \State $\alpha_1 \gets 1$, $\alpha_k \gets 0$ for all $k >1$     \Comment{reset BOCPD arrays}
    \State $\beta_1 \gets 1$, $\beta_k \gets 0$ for all $k >1$
    \State $\nu_1 \gets 1$, $\nu_k \gets 0$ for all $k >1$
    \State $\mathcal{I}_1 \gets 1$, $\mathcal{I}_k \gets 0$ for all $k >1$
\ElsIf{$t= T$}
    \State $E \gets (\vx, t, {\Sigma_\text{cuml}} / {n_\text{cuml}})$ \Comment{flush out any pending updates}
\EndIf
\EndFor
\EndFor\\
\Return $E$
\EndFunction
\end{algorithmic}
\end{algorithm}

\clearpage

\subsection{Spatiotemporal-Chunk Event Camera}
\label{sec:supp_chunk}

\begin{figure}[htp]
    \centering
    \includegraphics{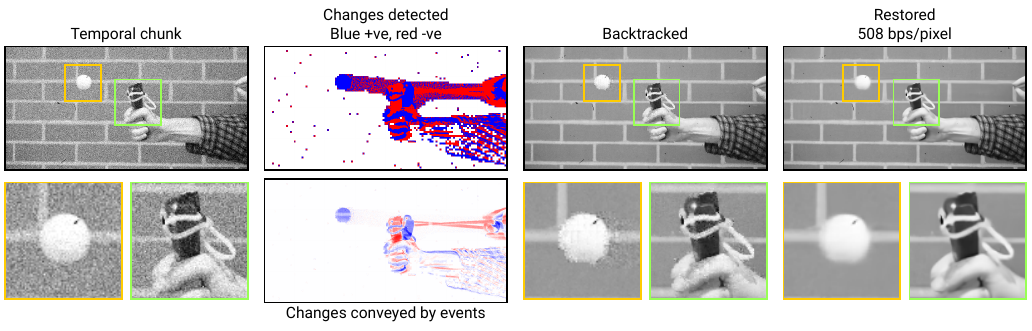}
    \vspace{-0.25in}
    \caption{\textbf{Intermediate outputs and recovered intensity from the spatiotemporal chunk method.} \textit{(left to right)} We first accumulate photon measurements into temporal chunks of, \eg, 32 binary frames. We then apply change detection on $4\times4$ spatial patches. Applying a restoration model to the backtracked outputs removes patch-boundary artifacts (\eg, patches that recently triggered an event may appear noisier than their neighbors).}
    \label{fig:patch_method_samples}
\end{figure}

\cref{alg:chunk-based} provides a formal description of the spatiotemporal-chunk event camera. Recall from \cref{sec:chunk_events} that this method generates an event whenever
\begin{equation}
    \norm{\mathbf{P}\paren{\tilde{\bm{\Phi}}_\text{chunk}(\vy, t) - \tilde{\bm{\Sigma}}_\text{patch}(\vy, t)}} \geq \tau.
\end{equation}
Let $p \times p$ be the patch size, with $q = p^2$ the number of pixels in a patch. We use a patch size of $4 \times 4$ in all our experiments (except for Fig.~4, where we use $8 \times 8$ for illustrative purposes). $\tilde{\bm{\Phi}}_\text{chunk}(\vy, t) \in \mathbb{R}^q$ is the normalized mean over a temporal chunk of $m$ binary frames; we use $m = 32$ throughout the paper. $\tilde{\bm{\Sigma}}_\text{patch}(\vy, t) \in \mathbb{R}^q$ is the normalized cumulative mean since the last event, excluding the current temporal chunk. $\bm{P} \in \mathbb{R}^{r \times q}$ is a feature matrix. We use $r = 16$ in our main experiments; for UltraPhase experiments, we reduce this to $r = 4$ due to on-chip memory constraints (see \cref{supp_sec:experiments-ultraphase}).

We normalize via
\begin{align}
    \tilde{\bm{\Phi}}_\text{chunk}(\vy, t) &= \bm{\Phi}_\text{chunk}(\vy, t) \oslash \mathbf{c} \\
    \tilde{\bm{\Sigma}}_\text{patch}(\vy, t) &= \bm{\Sigma}_\text{patch}(\vy, t) \oslash \mathbf{c},
\end{align}
where $\oslash$ indicates pointwise (Hadamard) division and $\bm{c}$ is given by
\begin{gather}
    \bm{c}(\vy, t) = \sqrt{\hat{\bm{p}}(\vy, t) \, (1 - \hat{\bm{p}}(\vy, t)) \frac{1}{m} \left( 1 + \frac{1}{n} \right)}, \\
    \hat{\bm{p}}(\vy, t) = \frac{n \bm{\Sigma}_\text{patch}(\vy, t) + \bm{\Phi}_\text{chunk}(\vy, t)}{n + 1},
\end{gather}
where $m$ is the temporal chunk size, $n$ is the number of temporal chunks comprising $\bm{\Sigma}_\text{patch}$, and the square root is pointwise. $\bm{c}^2$ is an empirical estimate of the variance in $\bm{\Phi}_\text{chunk} - \bm{\Sigma}_\text{patch}$ under a static assumption, in which case $\bm{\Phi}_\text{chunk}$ and $\bm{\Sigma}_\text{patch}$ are binomial random variables with the same probability. We use ghost sampling when computing $\hat{\bm{p}}$ to prevent numerical instability near $\hat{p} = 0$ and $\hat{p} = 1$; specifically, we add $8$ ghost Bernoulli measurements, $4$ of which are successes.

\paragraph{Training matrix $\bm{P}$.} We use the procedure outlined in \cref{alg:chunk-autodiff} to generate outputs that can be used to train the matrix $\bm{P}$ via backpropagation through time (BPTT). \cref{alg:chunk-autodiff} does not involve any branched control flow, and gives outputs identical to \cref{alg:chunk-based}. The event-generation decision is represented by a binary value $k$. On the forward pass, $k$ is computed using a Heaviside function. On the backward pass, we approximate the gradients of the Heaviside using a surrogate gradient approach. Specifically, we replace the Heaviside gradients with those of a logistic sigmoid.

Note that \cref{alg:chunk-autodiff} also includes an autodiff-compatible backtracking step, where we fill in the values of $\bm{B}$ in time-reverse order. Unlike an indexing operation, which is not differentiable with respect to its indices, this implementation is differentiable with respect to the values of $k$, which delimit the piecewise-constant segments of $\bm{B}$.

We train using interpolated videos from the XVFI dataset~\cite{Sim_2021_ICCV}. We interpolate from the native $1$~kHz frame rate to $16$~kHz using RIFE~\cite{huang2022rife}, then linearly interpolate over time by another factor of $8$ (for a total binary frame rate of $128$~kHz). We use the interpolated frames as the ground truth during training. To generate SPAD frames, we treat the ground truth values (in the range $[0, 1]$) as the Bernoulli photon-detection probability.

We initialize $\bm{P}$ with uniformly-distributed random values in the range $[-0.0125, 0.0125]$. We train for $20$ epochs, using vanilla SGD with a learning rate of $0.06$, reducing the learning rate by a factor of $5$ after $10$ epochs. Each epoch consists of $200$ batches, with each batch containing $64$ patch time series. Throughout training, we randomly vary the contrast threshold via uniform sampling in the range $[1 / 1.3, 1 / 0.7]$.

\begin{algorithm}
\caption{Spatiotemporal-Chunk Event Camera. We use $m = 32$. Typical values of $\tau$ are $[0.6, 1.4]$; however, this depends on $\bm{P}$, as there is a redundant degree of freedom between $\tau$ and $\bm{P}$}
\label{alg:chunk-based}
\begin{algorithmic}[1]
\Require {Patch-wise SPAD response, $\bm{\Phi}(\vy, t)$ \\
Contrast threshold, $\tau$\\
Feature matrix, $\mathbf{P}$\\
Temporal chunk size, $m$\\
Patch locations, $\mathcal{Y}$\\
Total bit-planes, $T$
}
\Ensure {Event-stream $E$ that contains packets of $(\vy, t, \text{patch-wise cumulative mean})$}
\Function{GenerateEvents}{$\bm{\Phi}(\vy, t)$, $\tau$, $\mathbf{P}$}
\For{$\vy \in \mathcal{Y}$}
\State Chunk-wise mean $\bm{\Phi}_\text{chunk} \gets \bm{0}$
\State Cumulative patch-wise mean $\bm{\Sigma}_\text{patch} \gets \bm{0}$
\State Cumulative counter $n \gets 1$
\For{$t \in \{1, 2, \ldots T\}$}
\State $\bm{\Phi}_\text{chunk} \gets \bm{\Phi}_\text{chunk} + \bm{\Phi}(\vy, t) / m$ \Comment{accumulate into chunk-wise mean}
\If{$t \in \{2m, 3m, \ldots\}$} \Comment{check for changes}
\State $\bm{\hat{p}} \gets (n \bm{\Sigma}_\text{patch} + \bm{\Phi}_\text{chunk}) / (n + 1)$ \Comment{empirical probability}
\State $\bm{c} \gets \sqrt{\bm{\hat{p}} (1 - \bm{\hat{p}}) (1 + 1 / n) / m}$ \Comment{normalization} \vspace{1em}
\If{$\norm{\mathbf{P}((\bm{\Phi}_\text{chunk} - \bm{\Sigma}_\text{patch}) \oslash \bm{c})}_2 \geq \tau$}. \Comment{change detected}
\State $E \gets (\vy, t, \bm{\Sigma}_\text{patch})$
\State $\bm{\Sigma}_\text{patch} \gets \bm{0}$
\State $n_\text{cumul} \gets 0$
\EndIf
\EndIf
\If{$t \in \{m, 2m, 3m, \ldots\}$}
\State $\bm{\Sigma}_\text{patch} \gets (n \bm{\Sigma}_\text{patch} + \bm{\Phi}_\text{chunk}) / (n + 1)$ \Comment{update cumulative mean}
\State $n \gets n + 1$
\State $\bm{\Phi}_\text{chunk} \gets \bm{0}$ \Comment{reset chunk-wise mean}
\EndIf
\EndFor
\EndFor\\
\Return $E$
\EndFunction
\end{algorithmic}
\end{algorithm}

\begin{algorithm}
\caption{Autodiff-Compatible Event Generation and Backtracking for Spatiotemporal-Chunk Event Camera. We treat SurrogateHeaviside as a Heaviside function on the forward pass, and a logistic sigmoid on the backward pass.}\label{alg:chunk-autodiff}
\begin{algorithmic}[1]
\Require {Patch-wise SPAD response, $\bm{\Phi}(\vy, t)$ \\
Contrast threshold, $\tau$\\
Feature matrix, $\mathbf{P}$\\
Temporal chunk size, $m$\\
Patch locations, $\mathcal{Y}$\\
Total bit-planes, $T$\\
Assume that $T \equiv 0 \pmod m$
}
\Ensure{Patch-wise $\bm{B}(\vy, t)$ that contains backtracked integrator values}
\Function{GenerateEventsAndSample}{$\bm{\Phi}(\vy, t)$, $\tau$, $\mathbf{P}$}
\For{$\vy \in \mathcal{Y}$}
\For{$i \in \{1, 2, \ldots, T / m\}$} \Comment{compute mean of each temporal chunk}
\State Chunk-wise mean for $i^\text{th}$ temporal chunk $\bm{\Phi}_\text{chunk}^{(i)} \gets \bm{0}$
\For{$1 \leq j \leq m$}
\State $\bm{\Phi}_\text{chunk}^{(i)} \gets \bm{\Phi}_\text{chunk}^{(i)} + \bm{\Phi}(\vy, m (i - 1) + j) / m$ \Comment{add bitplane to chunk-wise mean}
\EndFor
\EndFor
\State Cumulative patch-wise mean $\bm{\Sigma}_\text{patch}^{(1)} \gets \bm{\Phi}_\text{chunk}^{(1)}$
\State Cumulative counter $n \gets 1$
\For{$i \in \{2, 3, \ldots, T / m\}$} \Comment{generate events starting at the second chunk}
\State $\bm{\hat{p}} \gets \left( n \bm{\Sigma}_\text{patch}^{(i - 1)} + \bm{\Phi}_\text{chunk}^{(i)} \right) / (n + 1)$ \Comment{empirical probability}
\State $\bm{c} \gets \sqrt{\bm{\hat{p}} (1 - \bm{\hat{p}}) (1 + 1 / n) / m)}$ \Comment{normalization}
\State $k^{(i)} \gets \text{SurrogateHeaviside} \left( \norm{\mathbf{P}((\bm{\Phi}_\text{chunk}^{(i)} - \bm{\Sigma}_\text{patch}^{(i - 1)}) \oslash \bm{c})}_2 - \tau \right)$ \Comment{binary value indicating change detection}
\State $n \gets n \left( 1 - k^{(i)} \right) + 1$
\State $\bm{\Sigma}_\text{patch}^{(i)} \gets \bm{\Sigma}_\text{patch}^{(i - 1)} (n - 1) / n + \bm{\Phi}_\text{chunk}^{(i)} / n$ \Comment{update cumulative mean}
\EndFor
\State $\bm{B}(\vy, T / m) \gets \bm{\Sigma}_\text{patch}^{(T / m)}$
\For{$i \in \{T / m - 1, \ldots, 2, 1\}$}  \Comment{iterate backward}
\State $\bm{B}(\vy, i) \gets \left(1 - k^{(i)} \right) \bm{B}(\vy, i + 1) + k^{(i)} \bm{\Sigma}_\text{patch}^{(i)}$ \Comment{use next value of $\bm{B}$ if there is no event}
\EndFor
\EndFor\\
\Return $\bm{B}$
\EndFunction
\end{algorithmic}
\end{algorithm}

\clearpage

\subsection{Coded-Exposure Event Camera}
\label{sec:supp_coded}

\begin{figure}[htp]
    \centering
    \includegraphics[width=\textwidth]{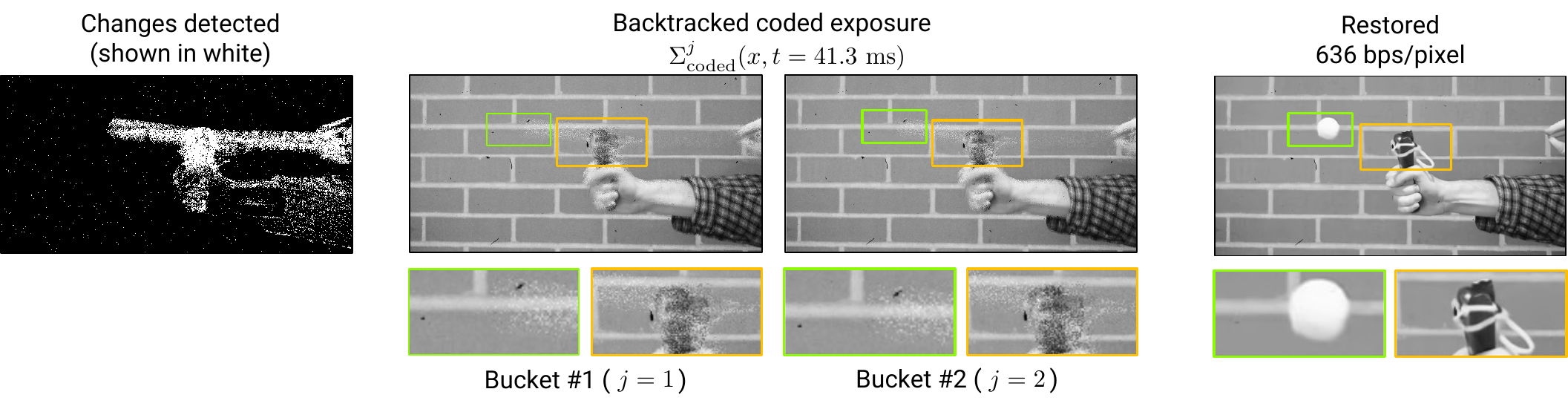}
    \vspace{-0.3in}
    \caption{\textbf{Intermediate outputs and recovered intensity from (two-bucket) coded-exposure events}. \textit{(left)} We show the changes detected using the confidence-interval test between two coded measurements that were computed across a temporal chunk of $1000$ binary frames. \textit{(middle)} The coded-exposure measurements, $\Sigma^j_\text{coded}$, differ predominantly in dynamic regions, while being statistically similar in static regions---this fact forms the basis of the change detector that we design for coded-exposure events. \textit{(right)} Reconstructions obtained using our video restoration model on a stack of pre-processed (pseudo-inverse step) and backtracked coded exposures.}
    \label{fig:coded-exposure-overview}
\end{figure}

\paragraph{Mask pattern (coding) details.} We choose $J \geq 2$ random binary mask sequences of length $N$ each, \ie, $\bm{C}^j \in \{0, 1\}^N$. Each of these $J$ sequences is non-overlapping, which is
\begin{equation}
    \bm{C}^j(n) \bm{C}^k(n) = 0 \;\forall\; j \neq k, \;\forall\; 1 \leq n \leq N.
\end{equation}
The motivation behind such a choice is to ensure that the pseudo-inverse step (Moore-Penrose inverse), which is applied to the coded measurements as a processing step, can be computed efficiently. With these constraints, the pseudo-inverse step involves multiplication (of the coded measurements) by a diagonal matrix, which can be carried out efficiently.

One way to construct such a mask sequence is by choosing $j^* \sim \text{Uniform}(1, J)$ at each sequence location $n \in {1, 2, ..., N}$, and then setting $\bm{C}^{j^*}(n) = 1$ and $\bm{C}^{j}(n) = 0$ for all other $j \neq j^*$. In other words, we pick a random ``bucket'' at each subframe index $n$.

For instance, when $J=2$, this amounts to picking to mask sequences, $\bm{C}^1, \bm{C}^2 \in \{0, 1\}^N$, such that
\begin{equation}
    \bm{C}^1(n) = 1 - \bm{C}^2(n) \;\forall\; 1\leq n \leq N.
\end{equation}
This choice corresponds to the ``coded two-bucket'' camera~\cite{Wei_2018_ECCV}. Thus, these random binary masks can be seen as a generalization of computing a single coded measurement to computing $J$ coded measurements~\cite{sundar_sodacam_2023}. 

Finally, we remark that we do not consider the case of $J=1$ here, since we implement coded exposures \textit{computationally} on single-photon sensors---hence, the ``complementary'' coded exposure (or the two-bucket measurement) is always readily available. In other words, there is no drastic compute or memory overhead of a (computational) two-bucket coded exposure over a single coded exposure. This would not be the case if an \textit{optical} setup (\eg, using digital micromirror devices, DMDs) were used to implement coded exposures---where using a two-bucket measurement would likely entail a beam splitter and a second DMD.

\paragraph{Pseudo-inverse step.} After backtracking, we can sample a $J$ bucket coded-exposure measurement at any time $t$. Of course, if the pixel is static, we only get $1$ static measurement at the pixel location---so we repeat static measurements $J$ times. Before applying our video restoration module, we perform a pseudo-inverse step that is derived from the linear forward model of $J$-bucket coded exposures and is similar to the pseudo-inverse pre-processing step adopted for single-bucket compressive captures~\cite{Yuan2022PnPADMM,Yuan2022ELPUnfolding}. This pre-processing step involves computing
\begin{equation}
    Y(\vx, n) = \sum_{j=1}^J \frac{1}{\sum_{m=1}^N \bm{C}^j(\vx, m)} \Sigma^j(\vx, t) \bm{C}^j(\vx, n),
\end{equation}
giving us $N$ subframes from the $J$ coded measurements.

To summarize, for coded-exposure events, we follow these steps:
\begin{equation}
    \text{Backtrack and sample} \rightarrow \text{Repeat static regions by $J\times$} \rightarrow \text{Pseudo-inverse pre-processing} \rightarrow \text{Video restoration}.
\end{equation}
We show intermediate and final outputs of (two-bucket) coded-exposure events on the slingshot sequence in \cref{fig:coded-exposure-overview}.

\begin{algorithm}
\caption{Coded-Exposure Event Camera. We typically choose $J$ to be $2$, $4$ or $8$ and $N$ to (correspondingly) be $8$, $16$ or $32$. Further, we set $\gamma$ in the range $1.8$--$3.5$, with smaller values resulting in more sensitive (frequent) change detection.}\label{alg:coded-exposure}
\begin{algorithmic}[1]
\Require {SPAD response, $\Phi(\vx, t)$ \\
Temporal extent of integrator, $T_\text{code}$\\
Number of buckets, $J$\\
Subframes per code, $N$\\
Wilson's score significance, $\gamma$\\
Pixel locations, $\mathcal{X}$\\
Total bit-planes, $T$ \\
Assume that $T_\text{code} \equiv 0 \pmod N$, $T \equiv 0 \pmod {T_\text{code}}$}
\Ensure {Event-stream $E$ that contains packets of $(\vx, t, J\text{ coded exposures})$}
\Function{GenerateEvents}{$\Phi(\vx, t)$, $\gamma$}
\For{$\vx \in \cx$}
\State $J$ binary mask sequences $\bm{C}^j \in \{0, 1\}^N$ such that:
\State \qquad $\sum_{j=1}^J \bm{C}^j(n) =1$  for all $1 \leq n \leq N$ \Comment{codes sum to one at each subframe index}
\State \qquad $\bm{C}^j(n) \bm{C}^k(n) = 0$ for all $j \neq k$ and for all $1 \leq n \leq N$  \Comment{codes are mutually orthogonal}
\State Cumulative sum, $\Sigma_\text{long} \gets 0$
\State Cumulative counter, $n_\text{long} \gets 0$
\For{$t_\text{code} \in \{1, \ldots, T / T_\text{code}\}$}
    \State Coded exposure $\Sigma^j_\text{coded} \gets 0$ for all $1\leq j \leq J$
    \For{$n \in \{1, \ldots, N\}$}
        \State Denote $t_\text{start} =  T_\text{code}\paren*{t_\text{code} + n / N}$
        \State Compute $\Phi_\text{avg}$: the average of $\Phi(\vx, t)$ in the interval $\left[t_\text{start}, t_\text{start}  + T_\text{code}/N \right)$
        \State $\Sigma^j_\text{coded}\gets \Sigma^j_\text{coded} + \bm{C}^j(n) \Phi_\text{avg} $, for all $1 \leq j \leq J$ \Comment{multiplex photon detections}
    \EndFor
    \If{$\Sigma^j_\text{coded} \in \text{conf}\paren*{T_\text{code} / J, \sum_s \Phi(\vx, s) / T_\text{code}}$ for all $j$} \Comment{binomial proportional test}
        \State $\Sigma_\text{long} \gets \sum_j \Sigma^j_\text{coded}$ \Comment{static region}
        \State $n_\text{long} \gets n_\text{long} + 1$
        \If{$t_\text{code} = T / T_\text{code}$}
            \State $E \gets (\vx, t_\text{code}, {\Sigma_\text{long}} / {n_\text{long}})$ \Comment{flush out any pending updates}
        \EndIf
    \Else
        \State $E \gets (\vx, t_\text{code}, \curly{\Sigma^j_\text{coded}}_{1\leq j \leq J} , {\Sigma_\text{long}} / {n_\text{long}})$
        \State $\Sigma_\text{long} \gets 0$ \Comment{reset long exposure}
        \State $n_\text{long} \gets 0$
    \EndIf
\EndFor
\EndFor\\
\Return $E$
\EndFunction
\end{algorithmic}
\end{algorithm}

\clearpage
\section{Extended Discussion}
\label{supp_sec:discussion}

\subsection{More Sophisticated Integrators}

In this work, we propose two integrators: adaptive exposures and coded exposures. There remains an extensive, unexplored space of alternate integrators. Both adaptive and coded exposures are linear projections of a pixel's photon detections over time. We could consider more general linear mappings, such as continuous-valued temporal codes. Further, we are not restricted to projections over time; it may also be advantageous to consider spatial projections, \eg, frequency-domain transforms. With these alternate projections, we might be able to reduce bitrates while maintaining reconstruction quality.

We have so far assumed that our objective is intensity reconstruction. However, there may be situations where we know that the final goal is a specific inference task. In such a scenario, we could design integrators that only encode information relevant to the task at hand; this may be significantly more bandwidth-efficient than transmitting a generic intensity encoding. This integrator could take the form of a learned module, \eg, a neural network, that operates near-sensor and computes a compressed, task-specific scene representation. We could train this module end-to-end with the downstream layers that perform final inference. The resulting system would involve a neural network that spans multiple compute devices, with an event-based communication layer in the middle. This setup somewhat resembles spiking neural networks and other event-based networks, although the goal with such methods is generally reduced computation costs (arising from sparse layer inputs) and not reduced bandwidth along a data-transfer interface.

\subsection{Entropy Coding and Quantization}

Generalized event cameras encode intensity levels, in the range $[0, 1]$, representing the photon-detection rate. In our experiments, we apply uniform quantization to the transmitted values; this keeps our comparisons straightforward and fair.

However, in a practical deployment, it may be more bandwidth-efficient to encode changes rather than values and apply entropy coding to the changes. When transmitting changes, the first event encodes a value in $[0, 1]$, and subsequent events encode a difference in $[-1, +1]$. This approach is functionally equivalent to transmitting levels or values, assuming the event camera correctly tracks quantization effects (to avoid drift). For natural scenes, the distribution of changes is non-uniform. The shape of this distribution depends in part on the change-detection algorithm; in general, we would expect changes near zero to be unlikely, as these would not trigger an event. We can apply entropy coding (\eg, Huffman coding) to exploit the nonuniformity in the distribution and achieve some additional compression. Entropy coding could give substantial bandwidth savings, albeit at the cost of some increased near-sensor computation.

In addition to entropy coding, we could apply non-uniform quantization, either to values or changes. For example, we could non-uniformly quantize values in $[0, 1]$ based on perceptual considerations (\eg, human sensitivity to intensity differences). With changes, we could exclude portions of the range $[-1, +1]$ based on the characteristics of the change detector; for example, with a fixed contrast threshold $\tau$, there is no need to represent changes in $(-\tau, \tau)$.

\subsection{Spatial Compression}

One advantage of patch-wise events (\eg, as described in Sec.~4.2) is that they permit some spatial compression. For example, we can apply JPEG block compression to the payload of an $8 \times 8$ patch-wise event. Such a change would bring our techniques more in line with existing video-compression algorithms, which compress in both space and time. The compression ratio we observe would depend on the sensor resolution; with higher resolution, we would expect image patches to be more uniform, and thus more easily compressible. Likewise, patches with more noise (\eg, due to a short adaptive integration window) would be more difficult to compress. Under ideal conditions, patch-wise compression might give us an additional $\tildeNice5\times$ reduction in bitrate. We leave this idea as a topic for future work.

\subsection{Alternate Sparse Formats}

As described in \cref{supp_sec:algos-overview}, we assume a COO event format $(\vx, t, \Lambda)$ when evaluating the bandwidth requirements of our methods. We could improve the sparse format to reduce bandwidth costs further. One such improvement would be to use an implicit time encoding. Specifically, on each binary frame, we would transmit a header $(t, n)$ indicating the current timestamp and the number of events on this frame. We would then send $n$ event packets $(\vx, \Lambda)$. If $n \gg 1$, this approach would virtually eliminate the overhead associated with transmitting values of $t$. Note, however, that this approach assumes we communicate events in time order; \ie, if $t_2 > t_1$, all events at time $t_1$ are sent before any events at time $t_2$.

To reduce the overhead associated with $\vx$, we could employ sparse matrix formats such as compressed sparse row (CSR) or compressed sparse column (CSC). These formats compress one of the two spatial coordinates (the row and column indices, respectively). The savings relative to COO would depend on the density of events on each frame, with denser events leading to more compression with CSR or CSC.

For pixel-wise methods (\ie, adaptive-EMA, Bayesian, and coded), another potential optimization is to adaptively group events into patch-wise packets. Assume we are using the COO format $(\vx, t, \Lambda)$. If a spatial patch contains many events, it may be more efficient to transmit them in a single packet, as this amortizes the overheads of $\vx$ and $t$. To implement this optimization, we would add an indicator bit $b$ to each event packet. If $b = 0$, then the packet should be treated as a pixel-wise event; if $b = 1$, it is a patch-wise event, meaning $\Lambda$ encodes integrator information for an entire patch. Within a patch-wise event we would use dense format for $\Lambda$, marking the pixels that triggered an event with a one-bit mask. Each patch would adaptively determine whether to encode its events pixel-wise or patch-wise. This decision would be based on the number of events, with the optimal threshold depending on the number of bits required for $\vx$, $t$, and $\Lambda$, as well as the patch size.

\subsection{Latency of the Adaptive Integrator}

One limitation of the adaptive integrator $\Sigma_\text{cumul}(\vx, T_1)$ is that it creates some latency in the intensity estimates. The estimate for the duration $[T_0, T_1]$---which the adaptive integrator computes as the mean of $\Phi(\vx, t)$ over $[T_0, T_1]$---is not known until $T_1$. Thus, there is a latency of $T_1 - t$ to obtain an estimate for $t \in [T_0, T_1]$. The expected delay depends on the frequency of events, with lower delay in more dynamic regions. The latency of $\Sigma_\text{cumul}(\vx, T_1)$ is not an issue for offline reconstruction and inference (which we assume throughout this paper). However, it may be of concern for certain real-time applications.

As a potential solution, we could introduce a second type of event, which we call an \textit{eager} event, in contrast to the \textit{change} events we have considered thus far. Assume again an event is generated at time $T_1$. In addition to transmitting $\Sigma_\text{cumul}(\vx, T_1)$ at that moment, we could send an eager event encoding a noisy estimate of $\Phi(\vx, T_1)$. We could then send periodic eager events encoding refinements to the value of the adaptive integrator. Assuming constant flux after $T_1$, the adaptive integrator converges to the true flux value as time passes; transmitting refinements would allow downstream components to leverage this improved estimate in the absence of a subsequent change event. We could consider various schedules for sending refinements---\eg, at exponentially increasing intervals, or at fixed intervals until some maximum time has passed. Note, however, that to keep the event camera's bandwidth coupled with (proportional to) the scene dynamics, there would need to be an upper bound to the number of eager refinement events, \ie, with each change event triggering at most $N$ refinements.

The solution described above would involve some additional bandwidth costs. We can thus imagine a generalized event camera that operates in two modes: ``online mode'' and ``offline mode,'' with eager events only being used in the online mode, where low latency is necessary.

\clearpage
\section{Restoration Model Details}
\label{supp_sec:restoration}

\subsection{Model Architecture}

While any video restoration model could be used, we choose the densely-connected residual network proposed in \citet{wang_2023_cvpr} (EfficientSCI) for our video restoration architecture---which was successful at restoring backtracked outputs of all of our generalized event cameras (adpative-EMA, adaptive-Bayesian, spatiotemporal chunk, and coded-exposure events). We also experimented with the spatial-temporal shift-based model of \citet{Li_2023_CVPR} (ShiftNet): while this architecture was successful at restoring three of our four proposed events, it did not succeed at restoring backtracked coded exposures. We attribute this to the fact that EfficientSCI was designed with video compressive sensing in mind, whereas ShiftNet is targeted for more general video-restoration tasks.

The memory cost of EfficientSCI scales with the number of input frames. Thus, we are constrained by the device memory in the number of frames we can reconstruct. For example, with $24$~GB of GPU memory and a resolution of $512 \times 256$, we can reconstruct about $96$ video frames---which can cover a temporal extent of $3000$--$9000$ binary frames ($30$--$90$ ms). Increasing the temporal extent requires sampling the backtracked cube at an increased temporal stride, which can lead to blurring and lower-quality results. One solution to this problem might be to employ a recurrent architecture with a fixed-size memory. However, forward-mode recurrent inference must be causal; \ie, the predicted frame at time $t$ may only consider backtracked samples from times before $t$. 
An efficient recurrent model may enable reconstructing videos at frame-rates faster than what we show in this work ($\tildeNice3000$ FPS)---indeed, our methods that operate the granularity of individual binary frames (\eg, \cref{sec:bitplane_events}) can, in principle, provide reconstructions at the frame-rate of SPAD photon-detections, \ie, $96.8$~kHz.

\subsection{Dataset}

We use $4403$ high-speed videos from the training split of the XVFI dataset~\cite{Sim_2021_ICCV} to train our restoration models. We temporally interpolate these $1000$ FPS to $16000$ FPS using
RIFE~\cite{huang2022rife} and treat each video frame, normalized between $0...1$ as the photon-detection probability. We then draw $6$ binary frames per video frame (as a per-pixel Bernoulli random variable, based on the photon-detection probabilities)---thereby giving us binary-valued responses at $96$ kHz. We remark that our dataset generation approach here (unlike in \cref{sec:rate-distortion-analysis,supp_sec:rate-distortion}) is not physically accurate, since we directly treat the video's values as photon-detection probabilities instead of average photon-arrival rates. We adopt this approach for its simplicity and note that the difference (between detection probabilities and Poisson rates) manifests as a tone-mapping operation by the SPAD's response function ($f(x) = 1-e^{-x}$). We did not see any generalization issues when applying our models trained this synthetic dataset to real SPAD data---in both ambient and low-light scenarios.

\subsection{Training Parameters}

All restoration models were trained until convergence ($40$--$60$ epochs) using the Adam optimizer \cite{kingma2014adam} and the mean squared error (MSE) loss objective. We used an initial learning rate of $10^{-5}$, which was decayed as per a cosine-annealed scheduler~\cite{loshchilov2016sgdr} to a minimum value of $10^{-8}$. When training the spatiotemporal chunk method, we randomly choose $\tau$ on each training iteration from a uniform distribution covering the range $[0.76, 1.43]$. We also clip the gradient norm to $1.0$ to resolve instability during training.

\clearpage
\section{Imaging Setup}
\label{supp_sec:cameras}
\begin{figure}[htp]
    \centering
    \includegraphics[width=0.7\textwidth]{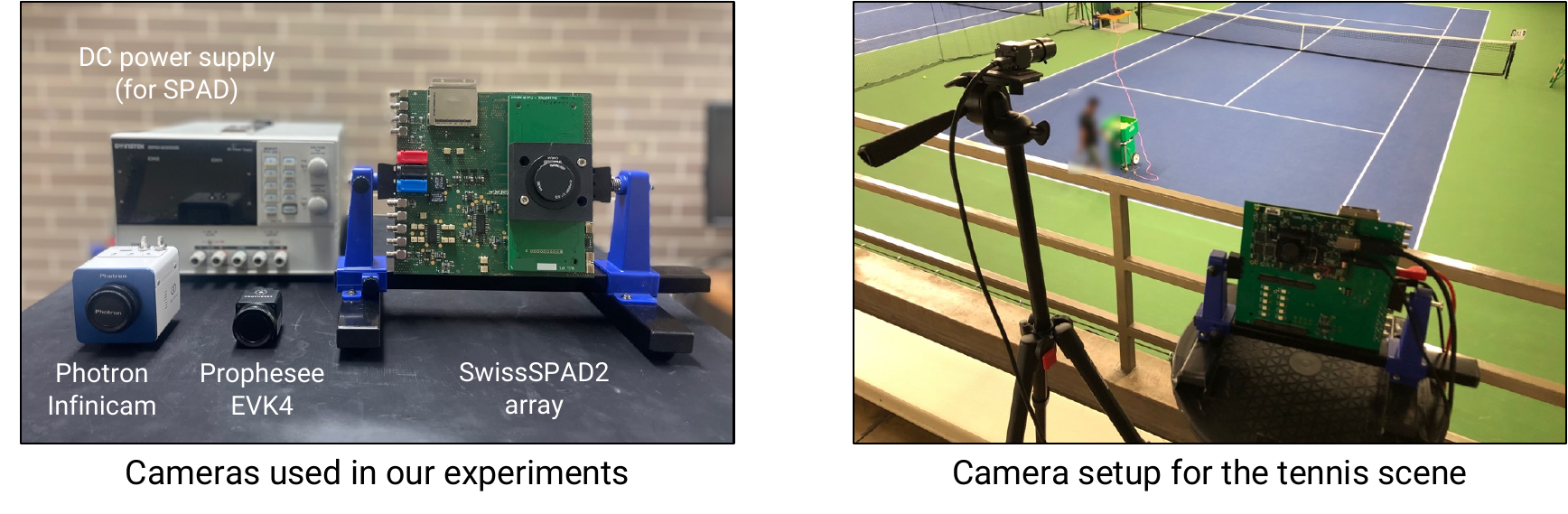}
    \vspace{-0.1in}
    \caption{\textbf{Camera setup for experimental acquisition}. \textit{(left)} Our setup comprises of the SwissSPAD2 array~\cite{ulku512512spad2019}, Prophesee EVK4 event camera and the Photron Infinicam high-speed camera. \textit{(right)} An example setup, which we used for capturing the tennis scenes shown in \cref{fig:first-figure,fig:plug_and_play}. We blur out the player for anonymity.}
    \label{fig:camera-setup}
\end{figure}

\cref{fig:camera-setup} shows the cameras we used for demonstrating the capabilities of generalized events: our algorithms process the outputs of the SwissSPAD2 array. We compare the performance of generalized events against the imaging capabilities of a commercial event camera (Prophesee EVK4) and a high-speed camera that is capable of real-time streaming (Photron Infinicam).

\subsection{Imager Specifications}

\paragraph{SwissSPAD2 Array.} The sensor has a resolution of $512\times 512$ pixels: however, we operate the device in its ``half-array mode'', meaning that only one subarray of resolution $512\times 256$ pixels is used. Each pixel has a pixel pitch of $16.4$ $\mu$m and a fill factor of less than $10\%$. The fill factor can be improved by the inclusion of microlens arrays which this prototype lacks.

The SPAD arrays features a certain number ($\tildeNice 5$\%) of ``hot pixels'': pixels whose dark count rate (DCR) is abnormally high, and as a result, almost always return $1$s. We calibrate a mask of hot pixels by using sensor responses that were taken in the dark. We find that hot pixels do not impact the generation of generalized events---thus, we inpaint hot pixels (either using the Telea algorithm~\cite{telea2004image} or nearest-neighbor replacement) using the calibrated mask after backtracking is performed, \ie, post-sensor readout.

\paragraph{Prophesee EVK4.} This commercially available event camera has a sensor resolution of $1280 \times 720$ pixels. Each pixel has a pixel pitch of $4.86$ $\mu$m and a fill factor of $>77\%$.

\paragraph{Photron Infinicam.} This high-speed camera provides a USB-C data interface, which is enabled by performing compression on the fly. The camera SDK does not provide access to parameters that control this online compression---as a result, we find that when imaging scenes where the camera is per-frame SNR limited, such as the indoor scene of \cref{fig:high_speed_recons}, read-noise induced artifacts translate to severe compression artifacts. The camera has a resolution of $1246 \times 1024$ pixels.

\subsection{Scene Acquisition Details}

We list the (focal lengths of C-mount) lenses used to capture each of the scenes shown in our main paper, along with the acquisition time of each scene (in terms of the number of binary frames captured by the SPAD at $96.8$~kHz).

\begin{itemize}
    \item Slingshot sequence in \cref{fig:first-figure}: $25$ mm lens, $4000$ binary frames.
    \item Tennis sequence in \cref{fig:first-figure}: $50$ mm lens, $3000$ binary frames.
    \item Vertical wheel in \cref{fig:first-figure}: $12$ mm lens, $3000$ binary frames.
    \item Nighttime traffic sequence in \cref{fig:first-figure}: $25$ mm lens, $3000$ binary frames.
    \item Dartboard sequence in \cref{fig:first-figure}: $16$ mm lens, $3000$ binary frames.
    \item Jack-in-the-box sequence in \cref{fig:telescoping}: $16$ mm lens, $4000$ binary frames.
    \item Rotating hole-saw bit sequence in \cref{fig:ema-bocpd}: $35$ mm lens, $6000$ binary frames.
    \item Casino roulette sequence in \cref{fig:patch_method}: $25$ mm lens, $4000$ binary frames.
    \item Stress ball sequence in \cref{fig:high_speed_recons}: $25$ mm lens, $4096$ binary frames.
    \item Nighttime traffic sequence in \cref{fig:low_light_recons}: $25$ mm lens, $8000$ binary frames.
    \item Tennis sequence in \cref{fig:plug_and_play}: $100$ mm lens, $8192$ binary frames.
\end{itemize}

\clearpage
\section{Extended Experiments}
\label{supp_sec:experiments}

In this supplementary note, we provide extended versions of our experiments in the main paper. We include detailed descriptions of our constructed baselines and make fine-grained comparisons across our proposed methods.

\subsection{Baseline Details}

\paragraph{EDI++ construction.} EDI~\cite{pantpami2022} is a hybrid event-plus-frame technique that can produce a series of sharp images from an event stream and a long exposure spanning the same time duration. There are two practical difficulties in implementing this method: (1) precise spatial- and temporal-alignment is needed between the frames and events, (2) differences in imaging modalities between conventional CMOS cameras and DVS event sensors. There is also a third (DVS specific) difficulty, which \citet{pantpami2022} overcome by solving an optimization problem (either across one pair of event streams and frames or across multiple such inputs): the contrast threshold in conventional event cameras is not necessarily known and can have inherent randomness~\cite{v2e_hu_2021_cvpr,Graca_2023_CVPR}. In contrast, when implementing EDI using SPAD-events and frames, we have none of these difficulties: frames and events are perfectly aligned (by construction), the same imaging modality (SPAD photon detection) is used to obtain both events and frames, and finally, the forward (or event generation) model is precisely controlled (no optimization problem has to be solved). Thus, SPAD-based EDI can be thought of as an ideal version of EDI. We further refine EDI outputs using a trained restoration model (with the same architecture used for video restoration of our generalized events), and call this idealized, refined version of EDI ``EDI++''.

\paragraph{8-bucket compressive sensing.} We use the multi-bucket video compressive sensing method described in \citet{sundar_sodacam_2023}, with $8$ buckets and spanning $32$ subframes---where the sum of $64$ binary frames comprises a single subframe. For restoring these coded $8$-bucket capture, we use EfficientSCI~\cite{wang_2023_cvpr}, while retaining the same densely-connected residual architecture that constitutes the video restoration model for the proposed generalized events.

\paragraph{Burst denoising baseline.} We perform burst denoising on a stack of $32$ short exposures---where the sum of $64$ binary frames constitutes one short exposure---using the align-and-merge technique of \citet{hasinoff_burst_2016}. After the merging step, we additionally perform BM4D denoising~\cite{maggioni2012nonlocal} (with $\sigma=0.02$, after binomial variance stabilization~\cite{yu2009variance}). Further, we find that BM4D applied directly to the stack of short exposures (with binomial variance stabilization) results in poor performance ($\tildeNice 28$ dB PSNR on the rate-distortion plot of \cref{fig:rate_distortion}).
\clearpage

\subsection{High-Speed Videography}

\begin{figure}[htp]
    \centering
    \includegraphics[width=\textwidth]{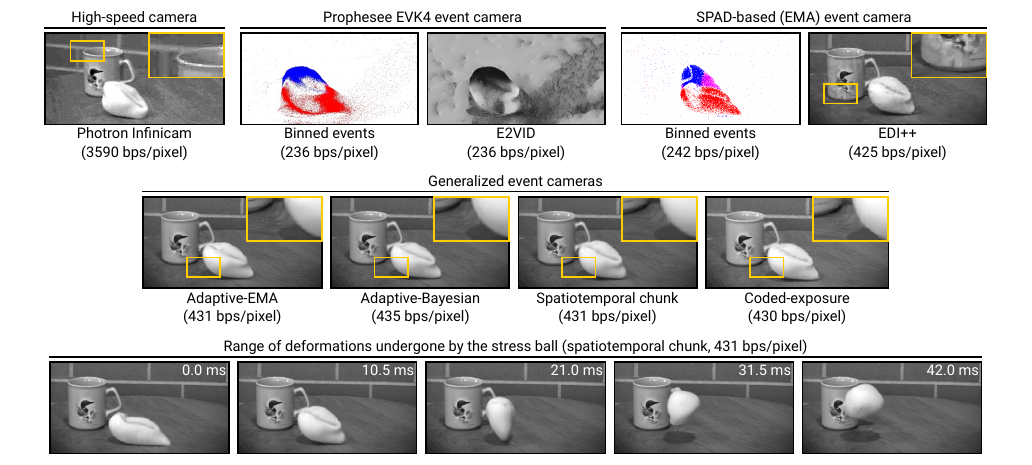}
    \caption{\textbf{Extended high-speed videography results.} In addition to the results shown in \cref{fig:high_speed_recons}, we show outputs obtained using our other event cameras, \textit{viz.} adaptive-EMA from \cref{sec:sigma-delta-instances}, adaptive-Bayesian from \cref{sec:bitplane_events}, and coded-exposure events from \cref{sec:coded_events}.}
    \label{fig:high_speed_recons_extended}
\end{figure}

\cref{fig:high_speed_recons_extended} contains extended results for high-speed videography on the stress ball sequence from \cref{fig:high_speed_recons}. We show that all of our generalized event cameras give comparable results for this scene.

\begin{figure}[htp]
    \centering
    \includegraphics[width=\textwidth]{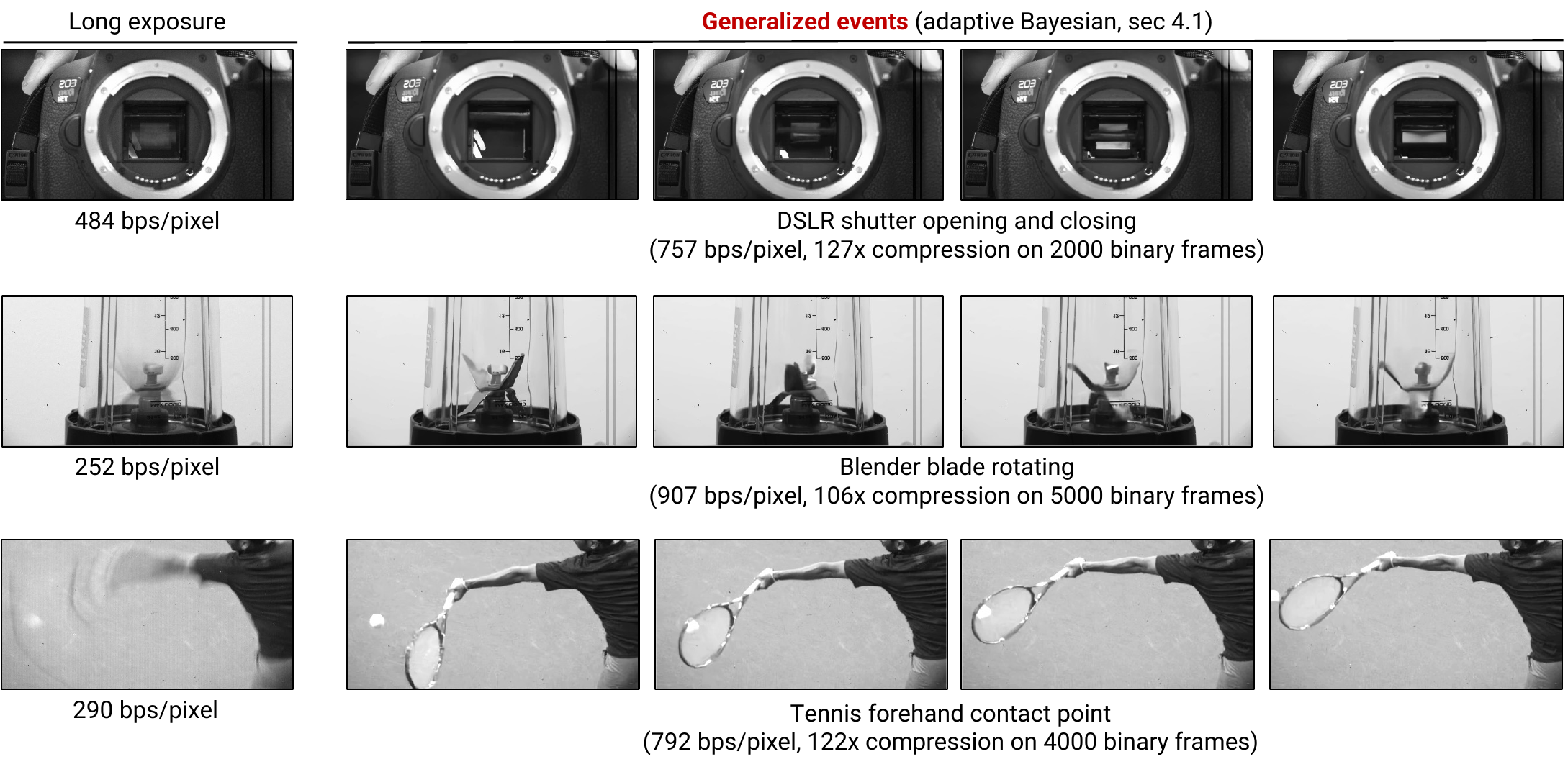}
    \vspace{-0.3in}
    \caption{\textbf{Additional high-speed videography results.} \textit{(left column)} We show a long exposure (average over all binary frames in the sequence) to depict the extent of motion in each sequence, \textit{(right)} and show $4$ video frames (uniformly separated in time across the capture duration) using our proposed method from \cref{sec:bitplane_events}. All videos were reconstructed at a frame rate of $3025$ FPS.}
    \label{fig:high_speed_recons_additional}
\end{figure}

We include additional results on three other sequences---a DSLR shutter in motion, the rotating blade of a blender, and a tennis forehand shot---in \cref{fig:high_speed_recons_additional}.

\clearpage

\subsection{Low-light Scenes}

\begin{figure}[htp]
    \centering
    \includegraphics[width=\textwidth]{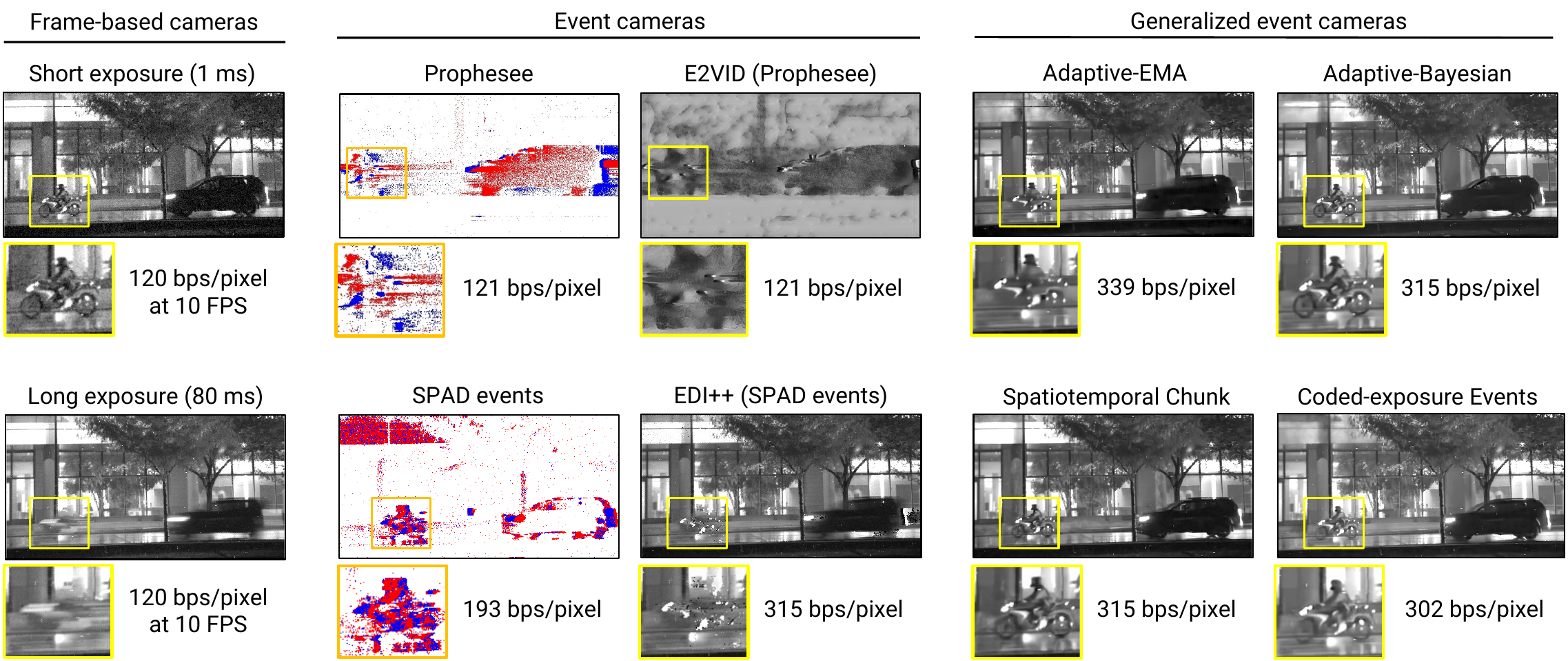}
    \caption{\textbf{Extended low-light results.} In addition to the results shown in \cref{fig:low_light_recons}, we show outputs obtained using our other event cameras, \textit{viz.} adaptive-EMA from \cref{sec:sigma-delta-instances}, spatiotemporal chunk from \cref{sec:chunk_events}, and coded-exposure events from \cref{sec:coded_events}.
    }
    \label{fig:low_light_recons_extended}
\end{figure}

\cref{fig:low_light_recons_extended} shows extended low-light results for the nighttime traffic scene in \cref{fig:low_light_recons}. The Bayesian, spatiotemporal chunk, and coded methods give better quality than adaptive-EMA, which fails to detect low-contrast changes (\eg, on the motorcycle wheel) due to its simplistic fixed-threshold change detector.

\begin{figure}[htp]
    \centering
    \includegraphics[width=\textwidth]{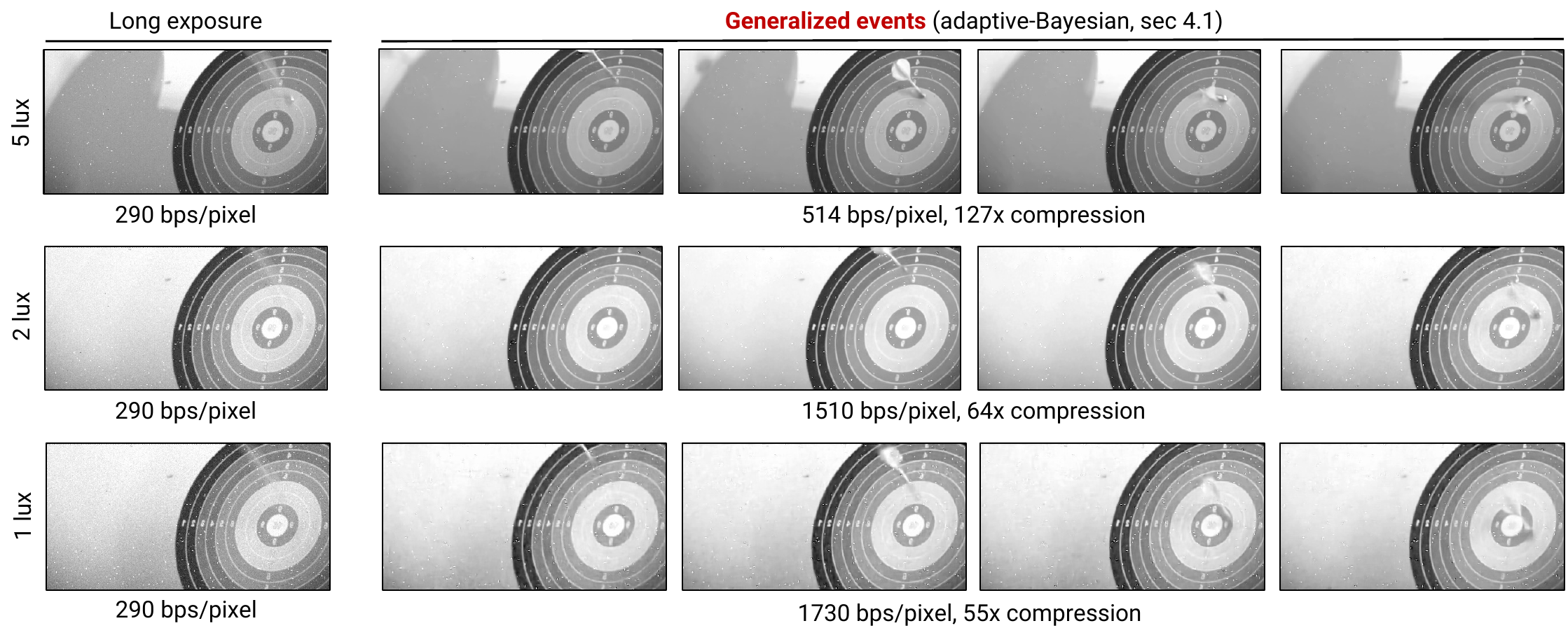}
    \vspace{-0.3in}
    \caption{\textbf{Additional low-light results.} We throw darts in the dark, at light levels of $5$, $2$ and $1$ lux (measured on the sensor side, \textit{top to bottom rows}). All three sequences span a duration of $41$ ms ($4000$ binary frames). Our restoration models are not trained on low-light sequences, despite this, we see reasonable low-light performance (\eg, at $5$ lux). At $1$ lux and $2$ lux, we see (somewhat graceful) degradation in our reconstructions. Lower the light level, harder it is to reliably distinguish scene motion from noise---which results in lower compression rates (and image quality) at lower lux values. Please zoom in to see details.}
    \label{fig:low_light_recons_additional}
\end{figure}

We provide an additional result on an indoor sequence shot in the dark (see \cref{fig:low_light_recons_additional}), which features lower (and controllable) light-levels---$1$, $2$ and $5$ lux measured using a light meter on the sensor side, as opposed to $7$ lux in \cref{fig:low_light_recons_extended}. We note that the low-light performance shown here could be improved upon with the inclusion of microlens arrays in the SPAD prototype, which could increase its fill factor (and in turn photon detection efficiency) from $10\%$ to $>40\%$.

\clearpage

\subsection{Scenes with Camera Motion}

In this subsection, we provide a qualitative analysis of the impact of ego-motion on the event-generation rate and the output-image quality. We consider three scenes with camera motion: the ``building'' sequence that features a significant amount of spatial structure and image detail (downtown buildings); the ``Ramanujan bust'' sequence, which is an indoor scene with a moderate amount of texture (from the bust and metal plaque); and a nighttime driving sequence where the SPAD was placed on the car's dashboard.

\begin{figure}[htp]
    \centering
    \includegraphics[width=\textwidth]{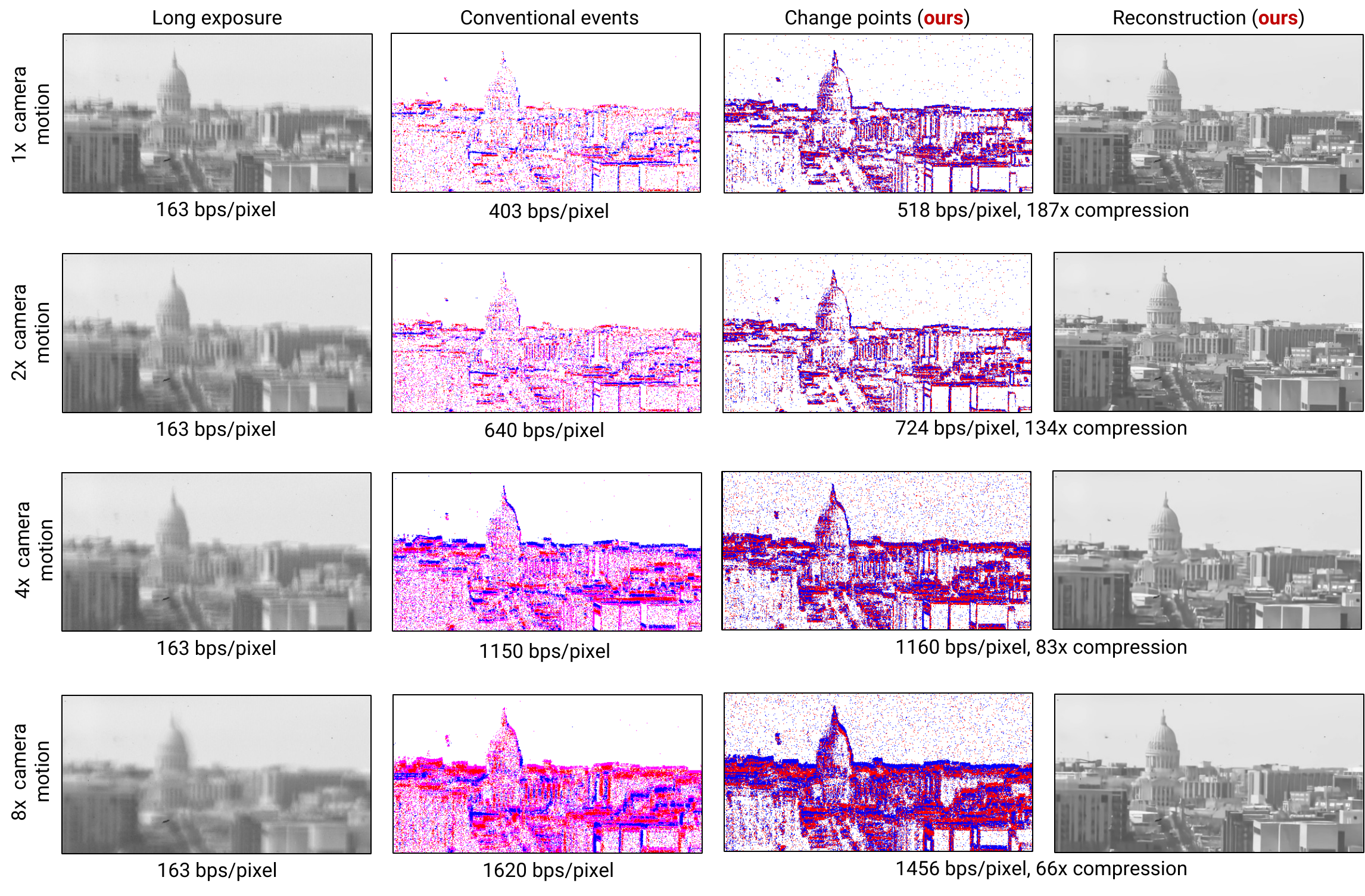}
    \caption{\textbf{Ego-motion results on the ``building'' sequence}. We run our generalized event technique on $8000$ binary frames in each row. In the second to fourth rows, we speed up photon-detections by processing only every $n$-th binary frame ($n=2,4,8$)---this exaggerates ego-motion. \textit{(left to right columns)} We depict the extent of motion using a long exposure across the same duration. We also show the changes detected by a (SPAD-based) event camera~\cite{sundar_sodacam_2023}. With increasing ego-motion, our techniques output more change points (as expected), but the reconstructions remain relatively sharp (there is some amount of blur induced), even with $8\times$ more motion.}
    \label{fig:camera-motion-capitol}
\end{figure}

\begin{figure}[htp]
    \centering
    \includegraphics[width=\textwidth]{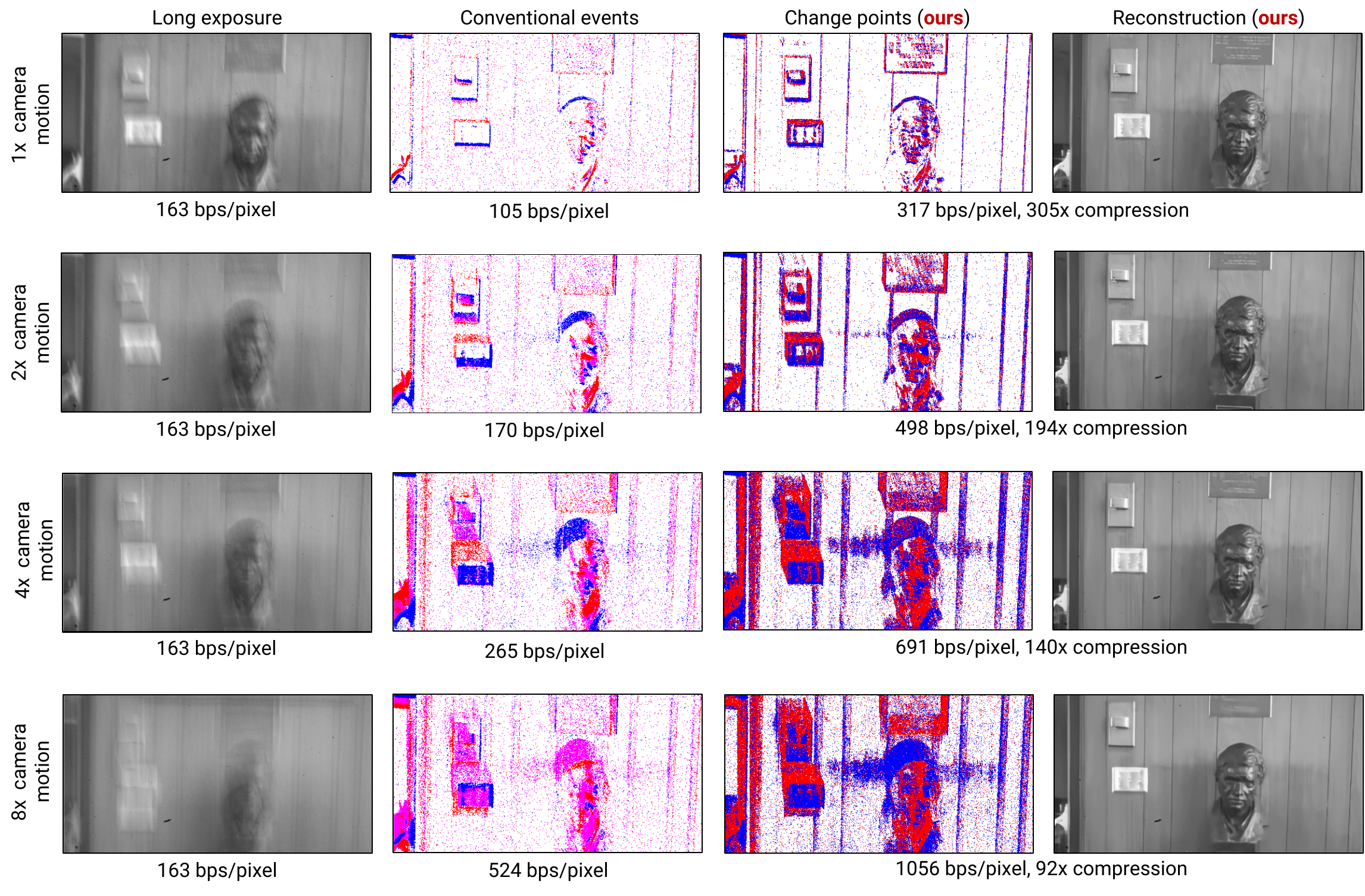}
    \caption{\textbf{Ego-motion results on the ``Ramanujan bust'' sequence}. The sequence comprises $8000$ binary frames, each acquired at a frame rate of $96.8$ kHz. Column and row descriptions are identical to \cref{fig:camera-motion-capitol}. This sequence has comparatively less texture than the building sequence, consequently, the blur in our output reconstructions (at $8\times$ the camera motion) is far less perceptible.}
    \label{fig:camera-motion-ramanujan}
\end{figure}

\begin{figure}[htp]
    \centering
    \includegraphics[width=\textwidth]{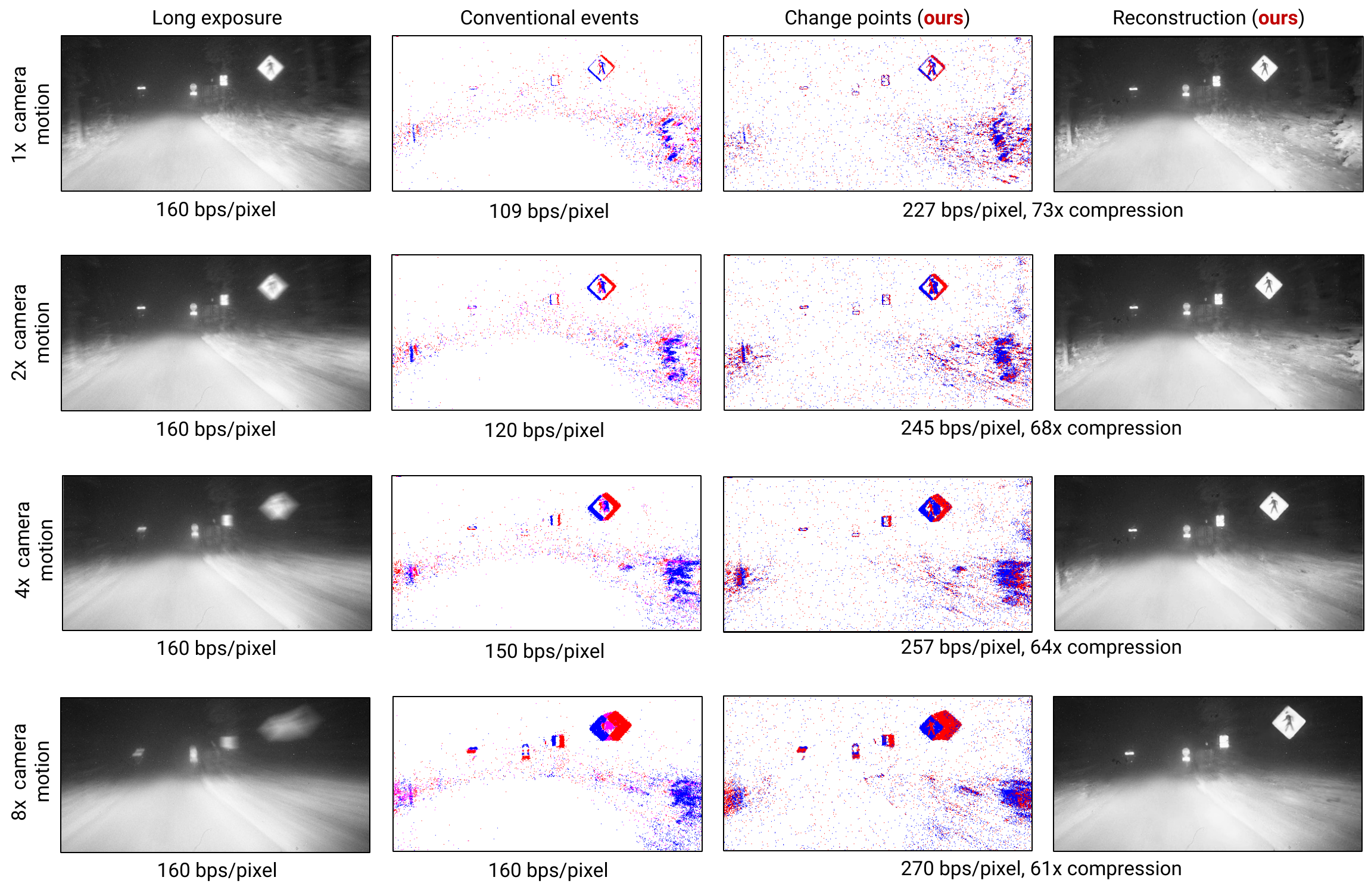}
    \caption{\textbf{Ego-motion results on the nighttime driving sequence}. Column and row descriptions are identical to \cref{fig:camera-motion-capitol}. The SPAD was operated at a lower speed in this sequence ($16.6$~kHz instead of $96.8$~kHz)---we report compression factors with respect to photon-detection readout at $16.6$~kHz. While this sequence also has lesser texture than \cref{fig:camera-motion-capitol}, the low-light conditions here make it more challenging. When ego-motion is exaggerated by $4\times$ (or higher, last two rows), we observe that details of the bushes are blurred out. However, the pedestrian crossing sign, which is has better contrast, is still recovered.}
    \label{fig:camera-motion-nighttime-drive}
\end{figure}

We show the change points and reconstructions obtained using one of our proposed generalized events (adaptive-Bayesian, \cref{sec:bitplane_events}) for these sequences in \cref{fig:camera-motion-capitol,fig:camera-motion-ramanujan,fig:camera-motion-nighttime-drive}. To bring out the effect of camera motion, we speed up the SPAD's output response (by skipping binary frames) by factors of $2\times$, $4\times$ and $8\times$. Thus, a $4\times$ sped up sequence sees $4\times$ as much camera motion. For additional context, we also show the extent of motion using a long exposure and the response of a (SPAD-based) event camera across the same duration.

Across all sequences, we observe that even with an exaggerated amount of camera motion (\eg, $4\times$ and $8\times$ sped-up sequences), we still see a significant amount of compression (reported with respect to raw photon readout) and a modest sensor readout (reported in bps/pixel).


\FloatBarrier

\subsection{Plug-and-Play Event Inference}

\begin{figure}[htp]
    \centering
    \includegraphics{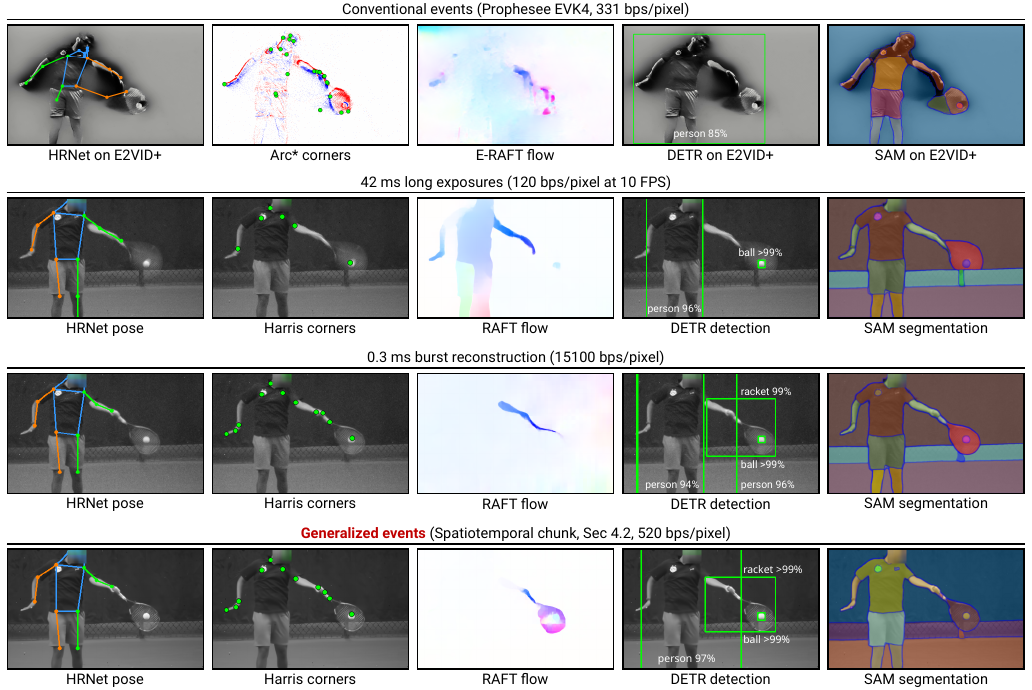}
    \tightcaption{\textbf{Expanded plug-and-play results.} In addition to the results shown in \cref{fig:plug_and_play}, we show results for long exposures (consisting of $4096$ binary frames) and burst reconstructions~\cite{ma_quanta_2020}.}
    \label{fig:plug_and_play_full}
\end{figure}

\paragraph{Expanded results.} \cref{fig:plug_and_play_full} shows an expanded set of plug-and-play inference results. This figure includes results for a $42$~ms long exposure ($4096$ binary frames, second row) and a burst reconstruction method~\cite{ma_quanta_2020}, run on consecutive $0.3$~ms short exposures ($32$ binary frames). The long exposure fails to capture fast-moving objects; there is significant blur in the arm, racket, and ball, causing inference to fail in these regions. The burst reconstruction gives good-quality inference in both static and dynamic regions; however, it requires a high readout bandwidth (about $30\times$ higher than our method).

\paragraph{Experiment details.} We manually trim the Prophesee outputs to a temporal extent corresponding to the SPAD capture (consisting of $8192$ binary frames). We run a Prophesee-provided E2VID model on these events to reconstruct a video. For our method and the burst reconstruction in \cref{fig:plug_and_play_full}, we run pose detection, corner detection, object detection, and segmentation on the reconstructed frame corresponding to the $2224^\text{th}$ binary frame. For the long exposure results in \cref{fig:plug_and_play_full}, we run these methods on the mean over binary frames $0$--$4095$. See below for optical flow extents. Below we provide additional details for some of the experiments in \cref{fig:plug_and_play,fig:plug_and_play_full}.
\begin{itemize}
    \item \textbf{HRNet pose:} We use the \href{https://github.com/open-mmlab/mmpose/blob/ffcfa39b073d1b20add2bd8611bf3b5fca9fe576/configs/body_2d_keypoint/topdown_heatmap/coco/td-hm_hrnet-w48_8xb32-210e_coco-256x192.py}{HRNet-W48} version of the model.
    \item \textbf{Harris corners:} We run a standard Harris corner detector with $\sigma = 4$.
    \item \textbf{RAFT flow:} For our method and the burst reconstruction in \cref{fig:plug_and_play_full}, we run RAFT between the reconstructions corresponding to the $2224^\text{th}$ and $2864^\text{th}$ binary frames. For the long exposure results in \cref{fig:plug_and_play_full}, we consider the interval between the $0$--$4095$ and $4096$--$8191$ exposures. We use the \href{https://pytorch.org/vision/main/models/generated/torchvision.models.optical_flow.raft_large.html#torchvision.models.optical_flow.raft_large}{RAFT-Large} version of the model.
    \item \textbf{DETR detection:} In most cases, we use a confidence threshold of $90\%$. We lower the threshold to $80\%$ for the E2VID reconstruction given the lack of high-confidence predictions. We use the \href{https://huggingface.co/facebook/detr-resnet-50}{ResNet-50} version of the model.
    \item \textbf{Arc* corners:} We run the algorithm on the entire event sequence. We temporally trim the predicted corners to a $0.8$~ms window around the target instant. In the figure, we show events within an $8$~ms window to provide visual context.
    \item \textbf{E-RAFT flow:} We prepossess events into a voxel grid with $30$ bins (divided into $2$ sub-durations), with a time span corresponding to that used in the RAFT experiments.
\end{itemize}

\clearpage

\subsection{Rate-Distortion Evaluation}
\label{supp_sec:rate-distortion}

\begin{figure}
    \centering
    \includegraphics{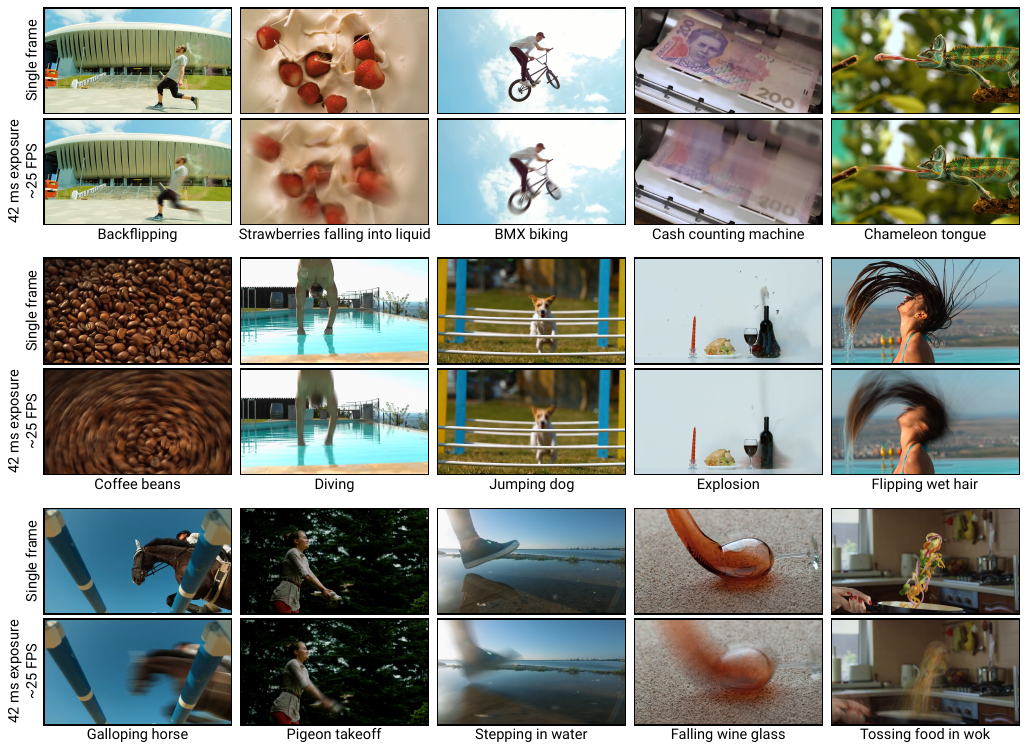}
    \vspace{-0.2in}
    \caption{\textbf{Thumbnails for YouTube-sourced videos.} Thumbnails from our \href{https://www.youtube.com/watch?v=A2eNVDGesYY}{YouTube-sourced evaluation dataset}. We show the central frame from each clip, along with a $42$~ms long exposure to illustrate the extent of motion.}
    \label{fig:youtube_thumbnails}
\end{figure}

\paragraph{Dataset details.} We source $15$ videos from YouTube that were captured by a Phantom Flex 4K camera at $1000$ FPS. \cref{fig:youtube_thumbnails} shows thumbnails depicting the scene content in each video, as well as a long exposure over $42$ ms that shows the extent of motion. We download these videos at a resolution of $854 \times 480$ pixels and further downsize (by $1.6\times$) and vertically crop them to the SPAD's resolution of $512 \times 256$ pixels.

We did not utilize the XVFI dataset~\cite{Sim_2021_ICCV}, which we used for training our video restoration models, for evaluating rate-distortion tradeoffs to prevent the possibility of data leakage. Further, we find that these YouTube-sourced videos have a more extreme range of motion than XVFI videos. 

\paragraph{Simulating SPAD responses.} To simulate SPAD photon detections from high-speed videos, we adopt the following steps:
\begin{enumerate}
    \item We interpolate videos at $1000$ FPS by $16\times$ using RIFE~\cite{huang2022rife} to $16000$ FPS.
    \item We treat the videos as sRGB-encoded frames, convert them to linear RGB images, and then grayscale.
    \item Next, we determine the average count of incident photo-electrons as
    \begin{equation}
        N(\vx, t) = \alpha \mathcal{I}(\vx, t) + d,
    \end{equation}
    where we $\mathcal{I}(\vx, t)$ represents a video frame, $d$ is number of spurious detections ($7.74\times 10^{-4}$ counts per binary frame, using values reported in \citet{ulku512512spad2019}). We choose $\alpha$ such that the average value of $\alpha \mathcal{I}(\vx, t)$ over each pixel location $\vx$ and time $t$ is $1$---we find that our ambient captures with the SwissSPAD2 have an average photon per pixel per binary frame (PPP) of $1$. 
    \item Finally, we draw binary frames with the probability of a $1$ given by
    \begin{equation}
        \Prob{\Phi(\vx, t) = 1} = 1 - e^{-N(\vx, t)}.
    \end{equation}
    From each frame at $16000$ FPS, we draw $6$ binary frames, thereby simulating photon detections at $96000$ Hz.
\end{enumerate}

\paragraph{Computing metrics.} We evaluate perceptual distortion using PSNR (computed from the average mean squared error across the entire video~\cite{kelecs2021computation}), SSIM (computed per-frame and averaged) and MS-SSIM (computed per-frame and averaged) metrics. Both metrics are converted with respect to linear values. Specifically, each of our generalized event cameras and baselines (EDI++, burst denoising, coded $8$-bucket) recovers the time-varying estimate of $1-e^{-N(\vx, t)}$, \ie, the probability of photon detection. Let us denote this by $\hat{p}(\vx, t)$. We can obtain linear estimates by computing
\begin{equation}
    \frac{1}{\alpha}\paren*{\log\paren*{\frac{1}{1 - \hat{p}(\vx, t)}} - d}.
\end{equation}

\paragraph{Baseline parameter sweeps.} We implement EDI++ with SPAD events, with an exponential decay of $0.95$, and sweep the threshold between $0.3$--$0.54$. The other baselines (burst denoising, long exposure, compressive sensing) are frame-based, and do not feature a tunable parameter that controls their readout rate.

\paragraph{Generalized-event parameter sweeps.} For adaptive-EMA, we set the exponential decay (of the EMA) to $0.95$, and sweep the threshold between $0.3$--$0.54$. For adaptive-Bayesian (\cref{sec:bitplane_events}), we retain the top-$3$ forecasters and vary the sensitivity $\gamma$ of BOCPD uniformly (on a logarithmic scale) between $10^{-7}$ and $10^{-2.2}$. For the spatiotemporal chunk method (\cref{sec:chunk_events}), we use a patch size of $4\times 4$ pixels, average $32$ binary frames per temporal chunk, and vary the threshold ($\tau$) of the change detector uniformly between $0.6$ to $1.28$. For coded-exposure events, we use a chunk size of $1024$ binary frames, with $4$ coded buckets (measurements per chunk) that multiplex $16$ subframes each---thus, each subframe consists of $1024/16 = 64$ binary frames. We vary the confidence level of Wilson's score (``$\alpha$'' parameter) uniformly between $1.2$ to $6.8$.

\begin{figure}[htp]
    \centering
    \includegraphics[width=\columnwidth]{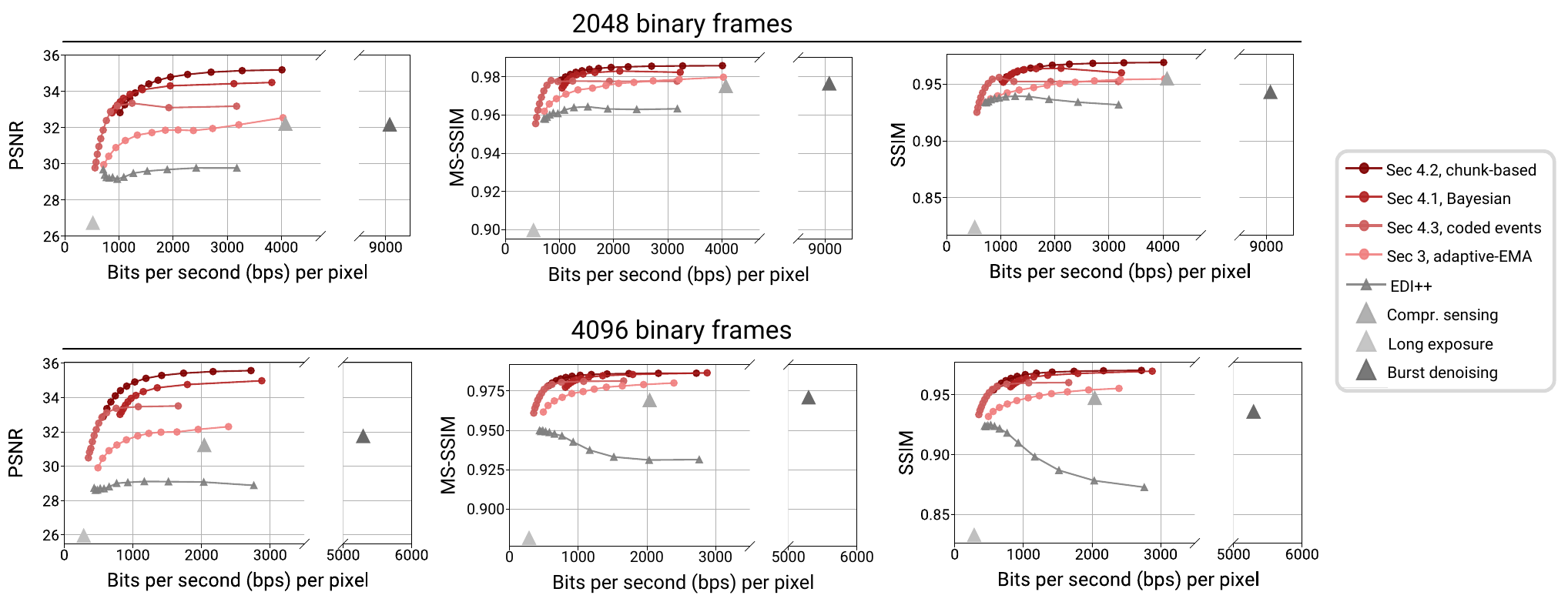}
    \tightcaption{\textbf{Extended rate-distortion evaluation.} \textit{(top)} Rate-distortion evaluations based on PSNR, SSIM and MS-SSIM metrics across $2048$ binary frames. \textit{(bottom)} When considering a longer temporal extent, \ie, $4096$ binary frames, we see that the readout rate at which a significant PSNR (or other metrics) drop-off is noticed becomes smaller---in other words, we obtain more compression of raw photon detections.
    }
    \label{fig:rate_distortion_extended}
\end{figure}

\paragraph{Extended results.} In addition to the PSNR-based rate-distortion plot shown in \cref{fig:rate_distortion}, we include evaluations based on SSIM and MS-SSIM metrics in \cref{fig:rate_distortion_extended} \textit{(top)}. Across metrics, we see that generalized event cameras provide a pronounced difference in performance over the considered baselines (shown in shades of gray). \textit{(bottom)} We include another evaluation on $4096$ binary frames (instead of $2048$). We observe that the readout rate, measured as bps/pixel, is lower across this extended duration---since the fixed readout costs of static (and less dynamic) regions is amortized over a longer duration, unlike frame-based cameras that involve fixed readout for all regions of an image. Finally, we remark that the performance gap between EDI++ and our techniques is further widened across this longer duration.

\FloatBarrier

\subsection{UltraPhase Experiments}
\label{supp_sec:experiments-ultraphase}

\paragraph{System description.} UltraPhase consists of $3 \times 6$ cores, each of which processes data from a $4 \times 4$ patch of SPAD pixels. The chip operates at a frequency of $0.42$~GHz, implying a maximum of  $4341$ instructions per binary frame at $96.8$~kHz. We additionally refer readers to \citet{ardelean2023computational} for a detailed description of the chip architecture.

\paragraph{Implementing event cameras on UltraPhase.} We implement our event camera designs in UltraPhase assembly code. Text files containing this code are included with this supplement (please see \path{assembly_adaptive_ema.txt}, 
\path{assembly_adaptive_bocpd.txt}, \path{assembly_spatiotemporal_chunk.txt} and \path{assembly_coded_exposure.txt}). 

\begin{figure}[htp]
    \centering
    \includegraphics{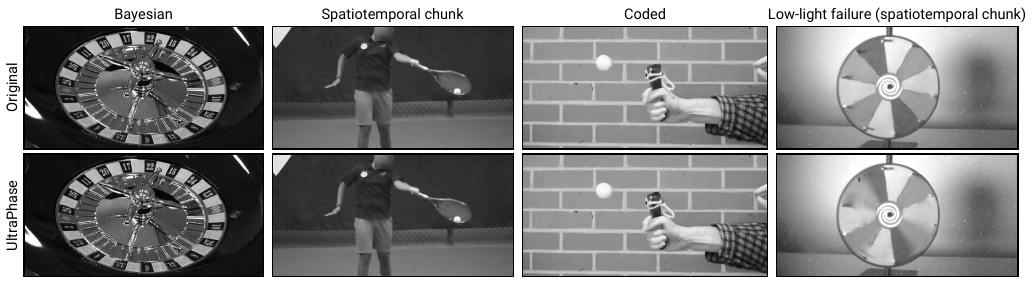}
    \tightcaption{\textbf{The effect of modifications for UltraPhase.} Some minor modifications are required to make our methods compatible with UltraPhase. As we observe in the first three columns, these modifications do not usually have any noticeable impact on the quality of the results; the modified methods \textit{(bottom)} give results that closely match the original methods \textit{(top)}. However, we do observe differences in low light, as seen in the rightmost column, which shows a scene captured at $0.3$~lux. In low light, a lack of noise-aware thresholding (\eg, the Wilson's bound for the coded method or the normalization step in the spatiotemporal chunk method) leads to less reliable change detections.}
    \label{fig:ultraphase_comparison}
\end{figure}

At this point, UltraPhase does not have native hardware for computing divisions---so we modify our methods to work avoid division operations. For the Bayesian method, we skip restarts and avoid division when computing the likelihood by multiplying arguments with their least common multiple---this is possible because the $\text{min}$ and $\text{argmax}$ operations carried out in BOCPD are invariant to the scale (of forecaster values). For the spatiotemporal chunk method, we skip the normalization step and use a $4 \times 16$, $8$-bit quantized feature matrix $\bm{P}$, in contrast to the $16 \times 16$ matrix we use in the rest of our experiments. For the coded-exposure method, we consider ($2$-bucket, $8$-subframe) masks; additionally, we replace Wilson's score by a fixed confidence interval (essentially amounting to a fixed threshold operation). In \cref{fig:ultraphase_comparison} we show comparisons between the modified and original methods; we see that, for most scenes, the above modifications do not significantly reduce the quality of the results.

\paragraph{Clock cycle measurements.} Due to circumstances beyond our control, we were unable to run our methods on a physical testbed system. We evaluate the runtime characteristics of our methods by assembling them for UltraPhase and measuring the number of compute cycles required to execute them. Given the deterministic nature of the digital hardware, the compute and memory requirements we measure in this evaluation are identical to those we would measure in physical hardware.

We confirm that all methods operate within the memory budget of the chip. In \cref{tab:ultraphase}, we show the measured clock cycles (the average per binary frame) for each method. In the case of branching, we assume the more computationally expensive branch is taken. Therefore, all compute values are an upper bound.

\begin{figure}[htp]
    \centering
    \includegraphics{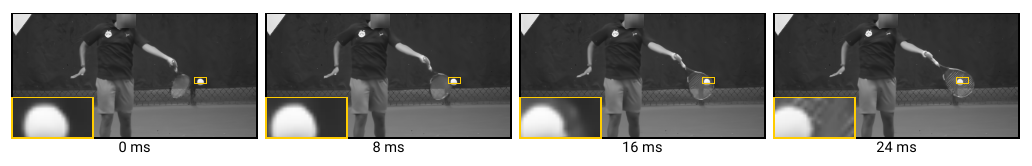}
    \tightcaption{\textbf{Sequence used for UltraPhase experiments.} This sequence covers $2500$ binary frames ($25.8$~ms). The cropped portion captures the edge of a swinging racket as it makes contact with a tennis ball.}
    \label{fig:ultraphase_crop}
\end{figure}

\paragraph{Readout estimation.} The chip readout depends on the dynamics of the scene. To estimate the readout, we run our methods on a $12 \times 24$ crop from the tennis sequence in \cref{fig:plug_and_play}, over $2500$ binary frames. We show frames from this crop in \cref{fig:ultraphase_crop}. We scale the measured readout values to units of kilobytes per second; see \cref{tab:ultraphase} for results.

\paragraph{Power estimation.} We estimate two components of power consumption: compute power and chip readout power. We base our analysis on \cite{sundar_sodacam_2023}. We assume the chip consumes $3.5$~picojoules per clock cycle spent executing an instruction, and that the chip expends $54$~nanowatts per kilobit of readout. \cref{tab:ultraphase} shows our estimated compute and readout power.

\begin{table*}[t]
    \centering
    \resizebox{\textwidth}{!}{\begin{tabular}{@{} l | c | c | c c c @{}}
        \toprule
        Method & Compute cycles & Bandwidth estimate (kB/s) & Compute power (W) & Readout power (W) & Total power (W) \\ 
        \midrule
        Raw photons          & $12$   & $3600$ & $5.43\times10^{-9}$ & $1.57\times10^{-3}$ & $1.57\times10^{-3}$ \\
        Adaptive-EMA         & $2367$ & $59.0$ & $2.11\times10^{-4}$ & $2.57\times10^{-5}$ & $2.37\times10^{-4}$ \\
        Bayesian             & $3095$ & $57.6$ & $3.62\times10^{-4}$ & $2.51\times10^{-5}$ & $3.87\times10^{-4}$ \\
        Spatiotemporal chunk & $380$  & $53.3$ & $5.45\times10^{-6}$ & $2.32\times10^{-5}$ & $2.87\times10^{-5}$ \\
        Coded                & $177$  & $33.1$ & $1.18\times10^{-6}$ & $1.44\times10^{-5}$ & $1.56\times10^{-5}$ \\
        \bottomrule
    \end{tabular}}
    \vspace{-0.1in}
    \caption{\textbf{UltraPhase results.} We measure the number of compute cycles (per binary frame) required to implement each of our methods, and estimate the readout bandwidth. Based on these values, we estimate the power required for on-chip computation and readout. All of our methods fit within the chip's computational budget of 4341 instructions per binary frame and give two orders of magnitude reduction in bandwidth compared to reading out raw photon detections. Due to these bandwidth reductions, our methods are also much more power efficient than raw photon readout.}
    \vspace{-0.15in}
    \label{tab:ultraphase}
\end{table*}

\clearpage

\end{document}